\documentclass[10pt,journal]{IEEEtran}

\usepackage{amssymb}
\usepackage[algoruled,vlined,linesnumbered]{algorithm2e}
\usepackage{comment}
\usepackage{amsthm}

\theoremstyle{definition}

\theoremstyle{remark}
\newtheorem{rem}{Remark}

\usepackage{diagbox}
\usepackage{bm}
\usepackage{amsmath}
\usepackage{float}
\usepackage{graphicx}
\usepackage{subfigure}
\usepackage{xcolor}
\usepackage{array}

\usepackage{hyperref}

\usepackage[colorinlistoftodos]{todonotes}

\makeatother
\ifCLASSOPTIONcompsoc
  \usepackage[nocompress]{cite}
\else
   \usepackage{cite}
 \fi
\begin{document}

\title{Meta-learning based Alternating Minimization Algorithm for Non-convex Optimization}
\author{Jing-Yuan~Xia,~\IEEEmembership{}
        Shengxi~Li,~\IEEEmembership{}
        Jun-Jie~Huang,~\IEEEmembership{}
        Zhixiong Yang,
        Imad~Jaimoukha,~\IEEEmembership{}
        and Deniz~Gunduz~\IEEEmembership{Fellow, IEEE}% <-this % stops a space

\IEEEcompsocitemizethanks{
\IEEEcompsocthanksitem Jing-Yuan Xia, Zhixiong Yang and Jun-Jie Huang are with College of Electronic Engineering and  College of Computer, National University of Defense Technology, Changsha, 410073, China.\protect\
\IEEEcompsocthanksitem Shengxi Li is with College of Electronic Engineering, Beihang University, Beijing, 100191,China.\protect\ 
\IEEEcompsocthanksitem Imad~Jaimoukha is with the Department of Electrical and Electronic Engineering, Imperial College London, SW7 2AZ, UK.\protect\
\IEEEcompsocthanksitem Deniz Gunduz is with the Department of Electrical and Electronic Engineering, Imperial College London, SW7 2AZ, UK and the Department of Engineering EnzoFerrari, University of Modena and Reggio Emilia (UNIMORE), Italy. \protect\

% % note need leading \protect in front of \\ to get a newline within \thanks as
% % \\ is fragile and will error, could use \hfil\break instead.

E-mail:(j.xia16, shengxi.li17, j.huang15,i.jaimouka,d.gunduz)@imperial.ac.uk.

\IEEEcompsocthanksitem *Jun-Jie Huang is the corresponding author.\protect\

This work is supported by National Natural Science Foundation of China, projects 61921001 and 62022091
% E-mail:(j.xia16, j.huang15,i.jaimouka)@imperial.ac.uk.
% \IEEEcompsocthanksitem X. Liu, is with College of Computer, National University of Defense Technology, Changsha, 410073, China.\protect\\ 
% E-mail: xinwangliu@nudt.edu.cn.

 }% <-this % stops an unwanted space

% \thanks{(Jun-Jie Huang is the corresponding author.)}
}

% \markboth{Journal of \LaTeX\ Class Files,~Vol.~14, No.~8, August~2015}%

% \markboth{Journal of \LaTeX\ Class Files,~Vol.~14, No.~8, August~2015}%

\IEEEtitleabstractindextext{%

\begin{abstract}
In this paper, we propose a novel solution for non-convex problems of multiple variables, especially for those typically solved by an alternating minimization (AM) strategy that splits the original optimization problem into a set of sub-problems corresponding to each variable, and then iteratively optimizes each sub-problem using a fixed updating rule. However, due to the intrinsic non-convexity of the original optimization problem, the optimization can be trapped into spurious local minimum even when each sub-problem can be optimally solved at each iteration. Meanwhile, learning-based approaches, such as deep unfolding algorithms, have gained popularity for non-convex optimization; however, they are highly limited by the availability of labelled data and insufficient explainability.
To tackle these issues, we propose a meta-learning based alternating minimization (MLAM) method, which aims to minimize a partial of the global losses over iterations instead of carrying minimization on each sub-problem, and it tends to learn an adaptive strategy to replace the handcrafted counterpart resulting in advance on superior performance. The proposed MLAM  maintains the original algorithmic principle, providing certain interpretability.   
We evaluate the proposed method on two representative problems, namely, bi-linear inverse problem: matrix completion, and  non-linear problem: Gaussian mixture models. The experimental results validate  the proposed approach outperforms AM-based methods. Our code is available at \href{https://github.com/XiaGroup/MF_MLAM}{https://github.com/XiaGroup/MF\_MLAM}.
\end{abstract}

}

\maketitle

\IEEEdisplaynontitleabstractindextext

\IEEEpeerreviewmaketitle

% \IEEEraisesectionheading{\section{Introduction}\label{sec:introduction}}
\section{Introduction}\label{sec:introduction}

\IEEEPARstart{I}{terative} minimization is one of the most widely used approaches in signal processing, machine learning and computer science.
% such as Alternating Direction Method of Multipliers (ADMM), Iterative Shrinkage-Thresholding Algorithm (ISTA), Alternating Least Square (ALS), and Expectation Maximization (EM). 
Typically, when dealing with multiple variables, these methods follow an alternating minimization (AM) based strategy that converts the original problem of multiple variables into an iterative minimization of a sequence of sub-problems corresponding to each variable while the rest of the variables are held fixed. 
However, due to the non-convexity of the problem, the obtained solutions do not necessarily converge to a global optimum, even when all the sub-problems are solved optimally at each iteration, see \cite{powell1973search} for examples. 
The major issue underlying the failure of AM when facing non-convexity is that a greedy and non-adaptive optimization rule is carried out when solving each sub-problem throughout the iterations. Therefore, it lacks sufficient adaptiveness and effectiveness in terms of handling local optimums.

Recent advances in deep learning have highlighted its success in obtaining promising results for non-convex optimization problems \cite{hu2020iterative,ren2020neural,bernstein2018signsgd,fan2018matrix}.
However, generic deep learning methods have limited generalization ability especially when test data is significantly different from the training data. This problem becomes more crucial in the ill-posed non-convex optimization tasks. The weak explainability of  deep neural network behavior also questions its applicability in certain scenarios.

\textit{Deep unfolding} \cite{gregor2010learning} as an alternative learning-based approach  has achieved significant success and popularity in solving various optimization problems. 
It improves model explainability by mapping a model-based iterative algorithm to a specific neural network architecture with learnable parameters. In this way, the mathematical principles of the original algorithm are maintained, thus leading to better generalization behavior \cite{you2021ista,fu2021deep,hu2020iterative,luo2020unfolding,pellaco2020deep,mai2021ghost,li2020deep}. 
We note that most of the deep unfolding algorithms are designed for solving  linear inverse problems and require supervised learning.
There are also deep unfolding algorithms for solving non-convex optimization problems.
Zhang \textit{et al.} \cite{luo2020unfolding} propose to unfold alternating optimization for blind image super-resolution, which is typically ill-posed and non-convex. 
In \cite{pellaco2020deep}, an unfolded WMMSE algorithm is proposed to estimate the parameter of the gradient descent step size for solving MISO beamforming problem, which is highly non-linear and non-convex. 
We can see from \cite{luo2020unfolding,pellaco2020deep}, when solving non-convex problems, deep unfolding algorithms either only retain the iterative framework and replace all components by deep networks \cite{luo2020unfolding}, or learn a minimum number of parameters but result in less effective performance \cite{pellaco2020deep}. 
There exists a trade off between achieving better performance with highly over-parameterized deep networks and retaining model explainability and generalization ability with minimum learnable parameters. This trade-off is also highly related to the amount of required labelled data and the request of prior knowledge.
Therefore, it is essential to design a new approach that enables us to carry learning-based neural network model with interpretable optimization-inspired behavior in an unsupervised learning way.

Meta-learning has witnessed increasing importance in terms of strong adaptation ability in solving new tasks \cite{ravi2016optimization,li2020continuous,liu2021gendet,chen2021multiagent,finn2017meta,li2016learning,xia2021meta}.
Unlike the standard supervised learning solutions, meta-learning does not focus on solving a specific task at hand but aims to learn domain-general knowledge in order to generate an adaptive solution for a series of new tasks. 
Typically, a meta-network collects domain-specific knowledge when solving each specific task, and then extracts domain-general knowledge across solving different tasks.
Most of the popular meta-learning algorithms, such as model-agnostic meta-learning (MAML) \cite{finn2017model}, { metric-based meta-learning \cite{li2020boosting}} and learning-to-learn \cite{andrychowicz2016learning}, share a hierarchical optimization structure that is composed of \emph{inner} and \emph{outer} procedures: the meta-network performs as an optimizer to solve specific tasks at inner procedure and the parameters of meta-network are then updated through the outer procedure. We note that the AM-based iterative method can also be regarded as sharing this bi-level optimization framework. However,  the optimization strategy in an AM-based algorithms is commonly frozen and the inner optimization behavior is independent to the outer alternating procedure.
Therefore, we are inspired to propose a meta-learning based alternating minimization (MLAM) algorithm for solving non-convex problems, which will enable inner optimization to be continuously updated with respect to the mutual knowledge extracted over outer steps.

The proposed MLAM algorithm is composed of two-level meta-learning, namely, the \textit{upper-level} and \textit{ground-level} meta learning. The upper-level meta-learning learns on a set of non-convex problems and aims to enhance the adaptability towards new problems. In contrast, the ground-level meta-learning learns on a sequence of sub-problems within each non-convex problem from the upper-level; and therefore, aims to find an adaptive and versatile algorithm for the sequential sub-problems.
The overview structure of the proposed MLAM method is shown in Fig.\ref{fig:MLAM-two-level}. 

Specifically, the upper-level meta-learning learns to leverage the optimization experiences on a series of problems, 
while the learned algorithm in the ground-level meta-learning maintains the original inner-and-outer iterative structure as well as the algorithmic principles, but replaces the frozen and handcrafted algorithmic rule by a dynamic and adaptive meta-learned rule.
In other words, it aims to learn an optimization strategy that is able to provide a "bird's-eye view" of the mutual knowledge extracted across outer loops for those sub-problems being optimized in the inner loops. Therefore,  the learned strategy does not optimize each sub-problem locally and exhaustively through minimization; instead,  it optimizes them by incorporating the global loss information with superior adaptability. 
% Besides, along with the upper-level meta-learning, the adaptability of the learned algorithm is further enhanced to enable good convergence and fast adaption on new non-convex problem instead of over-fitting to specific one. 
Moreover, the proposed MLAM algorithm is able to solve optimization problems in an unsupervised manner.
As a result, the proposed MLAM algorithm achieves better performance in  non-convex problems, while {requiring} less (and even no) labelled data for training.

The main contributions of this paper are mainly three-fold:
\begin{itemize}
\item The core contribution is the proposed meta-learning based alternating minimization (MLAM) approach for non-convex optimization problems. In an unsupervised manner, MLAM achieves a less-greedy and adaptive optimization strategy to learn a non-monotonic algorithm for solving non-convex optimization problems. 

\item The proposed MLAM takes a step further towards enhanced interpretability.
The algorithmic principles of the original model-based iterative algorithm is fully maintained without the need to replace iterative operations with black-box deep neural networks.

\item With extensive simulations, we have validated that the proposed MLAM algorithm achieves promising performances on the challenging problems of matrix completion and Gaussian mixture model (GMM). It is able to effectively solve these extremely difficult non-convex problems even when the traditional approaches fail.  
\end{itemize}
The rest of this paper is organized  as  follows. Section II gives the background and a brief review of previous approaches. Section III introduces our proposed MLAM approach and presents an LSTM-based MLAM method. Section IV illustrates two representative applications of our MLAM method. Section V provides simulation results and Section VI concludes the paper.

\section{Relevant Prior Work}
In this section, we will first briefly introduce the general problem formulation for multi-variable non-convex optimizations and the solution approaches.
Then, we will demonstrate our motivation of proposing  MLAM and  review the relevant meta-learning approaches.   

Non-convex optimization problems that involve more than one variable are of great practical importance, but are often difficult to be well accommodated.  The underlying relationship between variables can be linear (e.g., product, convolution) or non-linear (e.g., logarithmic operation, exponential kernel). For illustration convenience, we consider a general optimization formulation over an intersection of two variables in the matrix form, which can be expressed in the form of: 
\begin{equation}
    (\hat{\bm{W}},\hat{\bm{X}}) =  \underset{{(\bm{W},\bm{X})\in\mathcal{W}\times\mathcal{X}}}{\arg\min}F(\bm{W},\bm{X}),
    \label{overall problem}
\end{equation}
where  $F:\mathcal{W}\times\mathcal{X}\rightarrow\mathbb{R}$ is a non-convex function that describes the mapping between the observations $\bm{Y}=F(\bm{W},\bm{X})$ and two variables $\bm{W} \in \mathcal{W}$ and $\bm{X} \in \mathcal{X}$.

\subsection{Model-based Solutions}
The model-based iterative algorithms \cite{byrne2011alternating,ha2017alternating,niesen2007adaptive,sun2019iteratively,goldfarb2013fast,guminov2019accelerated,zhang2020robust,sui2020online,sui2020dynamic} typically solve (\ref{overall problem}) by adopting an AM-based strategy. The basic idea is to sequentially optimize a sub-problem corresponding to each variable whilst keeping the other variable fixed. That is, starting from an arbitrary initialization $\bm{W}_0\in\mathcal{W}$, the AM-based algorithm sequentially solves two sub-problems at the $t$-th iteration via:
\begin{equation}
    \begin{aligned}
    &\bm{X}_t = \arg\min_{\bm{X}\in\mathcal{X}}f_{\bm{W}_{t-1}} (\bm{X}),\\  
    &\bm{W}_t = \arg\min_{\bm{W}\in\mathcal{W}}f_{\bm{X}_{t}} (\bm{W}), \\
    \end{aligned}
    \label{eq:AM-sub}
\end{equation}
where $f_{\bm{W}_{t-1}}  (\bm{X})$ and $f_{\bm{X}_{t}} (\bm{W})$ are the functions corresponding to  $\bm{X}$ and $\bm{W}$, respectively, while fixing the other one to the value obtained in the previous iterations, i.e., $f_{\bm{W}_{t-1}}(\bm{X})=f(\bm{W}_{t-1},\bm{X})$ and $f_{\bm{X}_{t}}(\bm{W})=f(\bm{W},\bm{X}_{t})$. The solution of each sub-problem in (\ref{eq:AM-sub}) could be attained by a gradient-descent based iterative process, 
\begin{equation}
    \bm{X}_{i}=\bm{X}_{i-1}+ \phi\left(\{\bm{X}_{k}\}_{k=0}^{i-1}, \nabla f_{\bm{W}_{t-1}}(\bm{X}_{i-1}) \right),\label{sub-loop}
\end{equation}
where $\{\bm{X}_{k}\}_{k=0}^{i-1}$ represent the historical values of the parameters for $i$ optimization steps, $\nabla f_{\bm{W}_{t-1}}(\bm{X}_{i-1})$ is the gradient of objective function on $\bm{X}_{i-1}$, and $\phi(\cdot)$ defines the variable updating rule of different algorithms.  Algorithms such as ISTA \cite{you2021ista} and WMMSE\cite{pellaco2020deep} formulate a closed-form solution when iteratively solving each sub-problem in (\ref{eq:AM-sub}), which is essentially equal to a first order stationary point obtained by gradient descent based methods as well.

Before proceeding further, we first introduce some concepts that will be used throughout this paper. We define the \textit{overall problem} as the optimization problem with objective function $F(\bm{W},\bm{X})$, which will be called the global loss function. We refer to the optimization problems over each of the variables as the \textit{sub-problem}, and their objective functions  $f_{\bm{W}_{t-1}}(\bm{X})$ and $f_{\bm{X}_{t}}(\bm{W})$ for $t\geq1$, as local loss functions. We define the alternative iterations over the sub-problems as the outer-loop, and the iterative iterations for solving each sub-problem as the inner-loop.

The AM-based algorithm attempts to solve the overall problem by sequentially minimizing the two sub-problems $f_{\bm{W}_{t-1}} (\bm{X})$ and $f_{\bm{X}_{t}} (\bm{W})$.
However, the AM-based algorithm does not necessarily converge to a global optimal solution. 
This could be due to two main reasons: 1) AM-based methods optimize over the local loss functions without fully utilizing the information from the global loss function, and 2) AM-based methods usually solve the local loss function greedily using the first order information, which may not necessarily lead to the best solution, that is, the global optima, in terms of the global loss function. 
% We will elaborate these two issues in the sequel.

Addressing these two issues of the AM is non-trivial. The key difficult is that the variable optimized rules of the model-based solutions are frozen during the iteration, with respect to a certain update function, i.e., the $\phi$ in (\ref{sub-loop}). Typically, $\phi$ is designed for optimizing each sub-problem greedily, as a result, is not expected to reach the global optimal solutions for the overall problem.

\subsection{Learning-based Solutions}
Different from the model-based approaches, the recent deep learning based methods typically require to train an over-parameterized deep neural networks in an end-to-end learning fashion with a large labelled dataset \cite{dong2014learning,dong2016accelerating,he2016deep,samek2017explainable,you2021ista,zhang2020deep}. During testing, the trained deep neural network is fed by the observations and directly outputs the estimated variables.
% Recalling the formulation in problem (\ref{overall problem}), the end-to-end deep learning model can be expressed as:
% \begin{equation}
%     (\hat{\bm{W}},\hat{\bm{X}})=\text{{\fontfamily{cmss}\selectfont DeepNet}}(\bm{Y}),
% \end{equation}
% where $\bm{Y}$ is the input observation data to the deep neural network and $(\hat{\bm{W}},\hat{\bm{X}})$ are the estimate variables obtained by the black-box deep learning network behavior. 
The performance of these methods is highly bounded by the training datasets; however, the ground-truth data are neither sufficient nor even exist in realistic non-convex tasks, such as the GMM problems. 
Another shortcoming that is common to these learning-based methods is the weak explainability of the end-to-end deep neural network behavior. 
% Specifically, though the basic principle behind the training process is explicit, the consequential rules of the generators or predictors are not yet interpretable, and thus the network is typically considered to be a black-box model \cite{samek2017explainable}. 

Different from the generic deep learning approaches, deep unfolding algorithms try to combine model-based and learning-based approaches. They have three major features: 1) They map the iterative optimization algorithm into a specific unfolded network architecture with trainable parameters; 
2) each layer in the deep unfolding network corresponds to one iteration of the original iterative algorithm, while the number of layers, especially the iterations, is frozen; 
3)similar to the other deep learning approaches, deep unfolding also requires pair-wise labelled data for training.

Mathematically, deep unfolding approaches follow the iterative framework as in equation (\ref{eq:AM-sub}), but replace the analytical minimization algorithm (or specific operators such as soft-thresholding or singular value thresholding) by neural networks in the form of:
\begin{equation}
    \begin{array}{l}
        \bm{X}_{t}=\text{{\fontfamily{cmss}\selectfont Layer}}_{t}^{X}(\bm{W}_{t-1}),\\
        \bm{W}_{t}=\text{{\fontfamily{cmss}\selectfont Layer}}_{t}^{W}(\bm{X}_t),
    \end{array}
    \label{unfolding-loop}   
\end{equation}
where the number of iterations is fixed with $t=T$. Hence the whole deep unfolded network is composed by $T$ layers, where each layer is composed of several operators that reflect the mathematical behavior of the original iterative algorithm. 

In \cite{zhang2020deep}, Zhang et al. propose a deep unfolded network for image super-resolution in which the solver for one variable is a generic deep network while the other one keeps consistent with the model-based solver. In deep alternating network \cite{luo2020unfolding}, two networks, referring to an estimator and a restorer, work as two solvers for the splitted sub-problems. Therefore, the whole unfolding algorithm alternates between two network operators. In a different way, the unfolded network in \cite{pellaco2020deep} almost retains the original model-based iterative algorithm to solve a non-linear problem, but takes a network-based generator to learn the hyper-parameters for gradient descent step size by unsupervised learning. However, its performance does not surpass the counterpart model-based algorithm.   

In summary, deep unfolding has made a step further towards better explainability; however, these approaches still perform as an end-to-end network behavior and mostly only enable interpretable alternating structure while replacing the original optimization-based algorithm with deep neural networks. Consequently, its interpretability is limited when applied to ill-posed non-convex problems, and it is based on a data-driven optimization strategy. Besides, most deep unfolding algorithms  require a large number of labelled data for supervised learning. Hence, the performance of learning-based approaches on solving ill-posed non-convex problems, especially those without sufficient labelled data, is still limited.

\begin{figure*}[htbp!]
    \centering
    \includegraphics[width=0.65\textwidth]{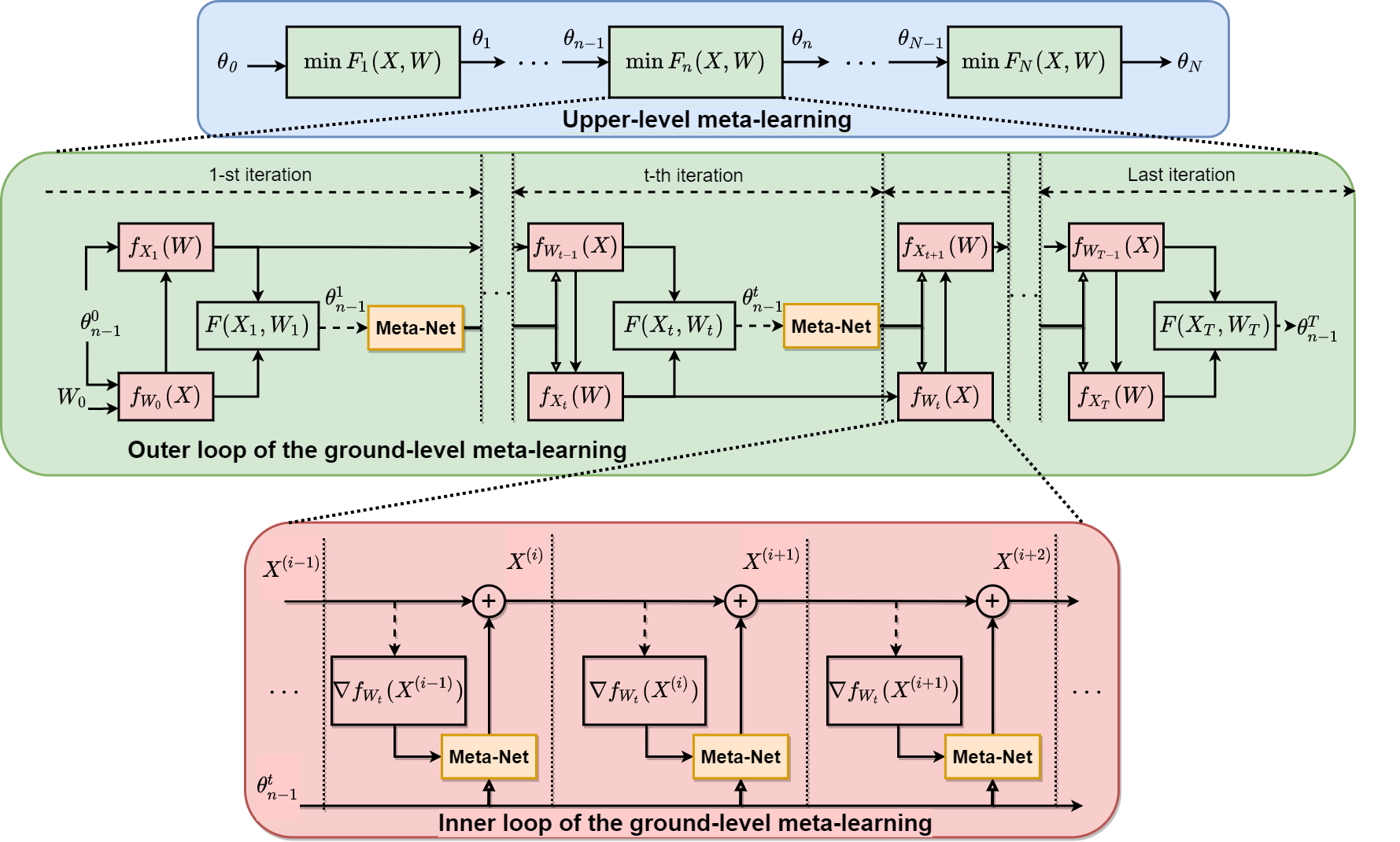}
    \caption{The overall structure of MLAM algorithm.
    The parameters of the applied meta network are denoted by $\theta$. The upper-level meta-learning is implemented on a set of non-convex problems $\{F_{n}(\bm{X},\bm{W})\}_{n=1}^N$ as shown in the upper row. $\theta$ is continuously updated across different problems. The ground-level meta-learning is applied to solve each non-convex problem $F_{n}(\bm{X},\bm{W})$ through an alternative outer loop between two inner loops corresponding to two variables, respectively, presented in the middle row. $\bm{\theta}$ is dynamically updated with respect to the global loss $F(\bm{X},\bm{W})$. The inner loop is depicted in the bottom row, where the variable is iteratively updated by the meta network. $\bm{\theta}$ is frozen at inner loop iterations. The dashed lines denote gradient operator occurs  and the solid line represents information flow along the edge.}
    \label{fig:MLAM-two-level}
\end{figure*}

\section{Meta-learning based Alternating Minimization (MLAM) Approach}
In this section, we introduce the proposed meta-learning based alternating minimization (MLMA) approach for solving optimization problems with multiple variables which are highly non-convex and ill-posed. As aforementioned, towards solving these problems, the existing model-based methods typically struggle, while on the other hand the lack of sufficient labelled training samples also restricts the performance of learning-based methods. Therefore, the key idea of our MLAM approach is to design a novel way that is not only capable of benefitting from both the optimization-based algorithmic principle and the superior performances by learning, but also surpass the AM-based strategy through meta-learning. 

\subsection{Overall Structure of the MLAM Approach}\label{sec:MLAM}
% Meta-learning for Alternating Optimization
The proposed MLAM approach consists of two levels of meta-learning. 
The upper-level meta-learning operates on a set of overall problems and 
the ground-level meta-learning optimizes the sequential sub-problems within an overall problem.
% , and the training on a set of overall problems refers to the upper-level meta-learning. 
Both upper-level and ground-level operations contain a bi-level optimization structure that is composed of outer and inner procedures. 
The outer loop of the upper-level meta-learning continuously updates the parameters of the $\text{{\fontfamily{cmss}\selectfont MetaNet}}$ $\bm{\theta}$ across different overall problems, and each inner loop equals to one ground-level meta-learning on solving an overall problem. 
For both the ground-level and the upper-level meta-learning processes, we denote note the inner loop index at superscripts and the outer loops index at subscripts.
We will then introduce the details of MLAM in a bottom-up manner. 
% We will demonstrate them respectively.
% In our MLAM method, there are two-level of meta-learning, 

\subsubsection{Upper-level Meta-learning}
The upper-level meta-learning is depicted in Fig. \ref{fig:MLAM-two-level} (the boxes in blue). It aims to extract a general knowledge of updating rules across different overall problems. 

Each overall problem is considered as a single task, and the whole learning process is accommodated on a set of tasks within the same dimension.
Therefore, the learnt algorithm is no longer designed for solving a single task, but a set of tasks $\{F_{i}(\bm{W},\bm{X})\}_{i=1}^{N}$. Although the proposed MLAM model follows the AM structure, it establishes a new bridge between the upper-level and ground-level meta-learning by replacing the frozen update functions with meta-networks. This allows for variable updated at the inner loops being guided by global loss information from the outer loops, thus achieving a global scope optimization.

\subsubsection{Ground-level Meta-learning}
The general structure of the ground-level meta-learning is shown in Fig. \ref{fig:MLAM-two-level} (the boxes in green and red). A ground-level meta-learning is performed by solving a specific overall problem $F_n(\bm{W}, \bm{X})$. This process contains an alternating optimization process on a sequence of sub-problems (boxes in green) which is named as the outer loop, and each sub-problem is solved by an iterative gradient descent based operation, which is defined as the inner loop (boxes in red). 

In the inner loop, the variable (\textit{e.g.,} $\bm{X}$ and $\bm{W}$) to be optimized is updated based on the iterative gradient descent process using a meta network and  the parameter of the $\text{{\fontfamily{cmss}\selectfont{MetaNet}}}$ is frozen.
Taking the optimization on $f_{\bm{W}_{t-1}}(\bm{X})$ as an example, the $i$-th step update on the variable $\bm{X}$ at the inner loop can be expressed in the following form:
\begin{equation}
  \bm{X}^{(i)}=\bm{X}^{(i-1)}+ \text{{\fontfamily{cmss}\selectfont MetaNet}}\left(\nabla f_{\bm{W}_{t-1}}(\bm{X}^{(i-1)}) \right),\label{mlam-inner-loop}   
\end{equation}
where $\text{{\fontfamily{cmss}\selectfont MetaNet}}(\cdot)$ is a neural network with learnable parameter $\bm{\theta}$, and performs as an optimizer for variable update.

MetaNet replaces the handcrafted update function $\phi(\cdot)$ in (\ref{sub-loop}) as a learnable and adaptive update function. 
Its parameter $\bm{\theta}$ is frozen at the inner loop and will be updated at the outer loop with respect to the global loss function. 
Specifically, different from the traditional AM-based algorithms, MLAM establishes an extra update cue for the parameters of $\text{{\fontfamily{cmss}\selectfont MetaNet}}$ at the outer loops, by minimizing the global loss $F(\bm{X},\bm{W})$. This enables that the gradients of $f_{\bm{W}}$ and $f_{\bm{X}}$ on variables $\bm{X}$ and $\bm{W}$ and the gradient of $F(\bm{X},\bm{W})$ on parameter $\bm{\theta}$ are integrated into one circulating system. In this way, the optimization behavior on each sub-problem is no longer independent to the others, all of which interact through the $\text{{\fontfamily{cmss}\selectfont MetaNet}}$. In Fig.\ref{fig:MLAM-two-level}, the dashed arrows at the outer loop of the ground-level meta-learning indicate the back-propagation of global loss to update the parameters $\bm{\theta}$ based on gradient descent. Taking the $t-th$ iteration as an instance, the parameter update is expressed as
\begin{equation}
    \bm{\theta}_{n-1}^t=\bm{\theta}_{n-1}^{t-1}+\alpha\cdot\Phi(\nabla_{\bm{\theta}_{n-1}^{t-1}} F(\bm{X}_1,\bm{W}_1)),
\end{equation}
where $\alpha$ is the learning rate, and $\Phi(\cdot)$ denotes the learning algorithms for network update (\textit{e.g.}, the Adam method \cite{kingma2014adam}).

Because the optimization landscapes for different sub-problems can be significantly different, 
% the learned update function tends to achieve a good and adaptive estimation for a sequence of sub-problems. 
MetaNet tends to extract a general knowledge on generating descent steps for variable updates across different sub-problems. 
% That is, the knowledge on generating descent steps for variables across different sub-problems can be successfully learned through the meta-learning process.  
Consequently, MetaNet learns to provide a superior and adaptive estimation for a sequence of sub-problems through this meta-learning process.

In Fig.\ref{fig:MLAM-two-level} the outer loop shows how the parameters at each iteration are updated. We define the update interval as the number of iterations for each parameter update, and in Fig.\ref{fig:MLAM-two-level} the update interval is set to 1. Generally speaking: the smaller the update interval the higher chances are updated, and the larger the update interval the more experience could be learned. The update interval of the parameters could be tuned. We will show more details in Section \ref{LSTM MLAM} and \ref{sec:matrix completion}.

\begin{rem}
The implementation of the two levels of meta-learning is indispensable. As the proposed MLAM learns in an unsupervised way, the training process of ground-level meta-learning could be also regarded as solving when training on one problem. Therefore, MLAM can be directly applied to a specific problem by only carrying on the ground-level meta-learning \cite{xia2021meta}. {In this paper, the proposed approach works with both of the two levels of the meta-learning.}
\end{rem}

{
The meta-learning implemented on the proposed MLAM approach mainly contributes to two aspects. (i) Recalling to the upper-level meta-learning, the training strategy of the MetaNet follows the meta-learning strategy, thus improving the adaptation of the learned parameters on solving different non-convex optimization problems $\{F_{i}(\bm{W},\bm{X})\}_{i=1}^{N}$. Specifically, the meta-learning behavior significantly enhance the capacity of solving different tasks by leveraging the experience of solving a series of $\{F_{i}(\bm{W},\bm{X})\}_{i=1}^{N}$.  (ii)The optimization performance of solving each specific task is dramatically improved by achieving a dynamic and adaptive gradient based updating rule through the implemented ground-level meta-learning. The solution strategy of solving each task $F_{i}(\bm{W},\bm{X})$ is meta-learned over the alternating iterations. In this way, the solution strategy is leveraged by the meta-learning behavior to allow the sub-problems $\{f_{\bm{W}_{t}}(\bm{X}),f_{\bm{W}_{t}}(\bm{X})\}_{t=1}^{T}$ to be solved in a less greedy but more effective way. Essentially, benefit of the meta-learning, the MLAM is capacity of leveraging the experience of solving a series of sub-problems $\{f_{\bm{W}_{t}}(\bm{X}),f_{\bm{W}_{t}}(\bm{X})\}_{t=1}^{T}$ to learn a gradient based strategy that provides better convergence on solving the current task $F(\bm{W},\bm{X})$.
}

\subsection{MLAM with LSTM-based MetaNet}\label{LSTM MLAM}
In this section, we introduce the implementation details of using recurrent neural networks (RNNs) as the $\text{{\fontfamily{cmss}\selectfont MetaNet}}$ in the proposed MLAM algorithm. Specifically, the Long Short-Term Memory ($\text{{\fontfamily{cmss}\selectfont LSTM}}$) network \cite{schmidhuber1997long} is adopted as the variable update function at the ground-level in the MLAM model. The proposed LSTM-MLAM updates the parameters of the MetaNet with respect to the accumulated global losses.

RNN has a sequentially processing chain structure to achieve the capacity of ``memory'' on sequential data. LSTM \cite{schmidhuber1997long} is one of the most well-known RNNs and is able to memorize and forget different sequential information. Memory is the most important feature of LSTM (RNN). It stores the status information of previous iterations and allows for the information to flow along the entire chain process. In this way, LSTM can integrate previous information with the current step input. 
Mathematically, the output of LSTM at the $i$-th iteration $\bm{H}^{(i)}$ is determined by the current gradient $\nabla f(\bm{x}_i)$ and the last cell state $\bm{C}^{(i)}$ in the following forms:
\begin{equation}\begin{array}{c}
     \bm{H}^{(i)}=\text{{\fontfamily{cmss}\selectfont LSTM}}(\nabla f(\bm{x}_{i}),\bm{C}^{(i)}, \bm{\theta}^{(i)}),  \\
     \bm{C}^{(i+1)}=\bm{z}_{f} \odot \bm{C}^{(i)}+ \bm{z}_{i} \odot \tilde{\bm{C}}^{(i-1)},
\end{array}\label{eq:cell state}
\end{equation}     
where $\text{{\fontfamily{cmss}\selectfont LSTM}}(\cdot,\cdot,\bm{\theta})$ denotes the LSTM network with parameters $\bm{\theta}$, $\odot$ denotes Hadamard Product, $\bm{z}_{f}$ and $\bm{z}_{i}$ are the vectors of intermediate conditions inside LSTM, and $\tilde{\bm{C}}^{(i-1)}=\nabla_{\bm{\theta}^{(i-1)}}\mathcal{L}^{(i)}$ is the candidate cell state, referring to the gradient of current loss $\mathcal{L}^{(i)}$ over the last parameters $\bm{\theta}^{(i-1)}$ in our problem. 

We adopt two LSTM networks $\text{{\fontfamily{cmss}\selectfont LSTM}}_{\bm{X}}$ and $\text{{\fontfamily{cmss}\selectfont LSTM}}_{\bm{W}}$ as {MetaNet} to generate variable update functions, recalling $\phi(\cdot)$ in (\ref{sub-loop}), for solving sub-problems corresponding to variables $\bm{X}$ and $\bm{W}$, respectively. We also denote $\bm{\theta}_{\bm{X}}$ and $\bm{\theta}_{\bm{W}}$ as the parameters of $\text{{\fontfamily{cmss}\selectfont LSTM}}_{\bm{X}}$ and $\text{{\fontfamily{cmss}\selectfont LSTM}}_{\bm{W}}$, and denote $\bm{C}_{\bm{X}}$ and $\bm{C}_{\bm{W}}$ as their cell states. The inputs of the LSTM are the gradient of local loss function and the sequential knowledge of variables, represented by cell state $\bm{C}$. Then the LSTM outputs the variable update term that integrates step size and direction together.
Denoting the inner loop update steps $i-1$ and $j-1$ at superscripts, and outer loops steps $t-1$ and $t$ at subscripts for each sub-problem, the variables are updated in the following forms:
\begin{equation}
    \begin{array}{cc}

        {}\bm{H}_{\bm{X}}^{(i-1)}
        =\text{{\fontfamily{cmss}\selectfont LSTM}}_{\bm{X}}\left(\nabla     f_{\bm{W}_{t-1}}(\bm{X}^{(i-1)}),~\bm{C}_{\bm{X}}^{(i-1)},~\theta_{\bm{X}} \right),
        \\\bm{X}^{(i)}=\bm{X}^{(i-1)}+\bm{H}_{\bm{X}}^{(i-1)},\\
        % \\ \bm{C}_{\bm{X}}^{(i)}=\bm{z}^{f} \odot \bm{C}_{\bm{X}}^{(i-1)}+ \bm{z}^{i} \odot\tilde{\bm{C}_{\bm{X}}}^{(i-1)}
    \end{array}
    \label{eq:x_update}
\end{equation}
and 
\begin{equation}
        \begin{array}{cc}
        {}\bm{H}_{\bm{W}}^{(j-1)}
        
      =\text{{\fontfamily{cmss}\selectfont LSTM}}_{\bm{W}}\left(\nabla f_{\bm{X}_{t}}(\bm{W}^{(j-1)}),~\bm{C}_{\bm{W}}^{(j-1)},~\theta_{\bm{W}}\right),
        \\\bm{W}^{(j)}=\bm{W}^{(j-1)}+\bm{H}_{\bm{W}}^{(j-1)}.\\
        %  \\ \bm{C}_{\bm{W}}^{(j)}=\bm{z}^{f} \odot \bm{C}_{\bm{W}}^{(j-1)}+ \bm{z}^{i} \odot\tilde{\bm{C}_{\bm{W}}}^{(j-1)}
    \end{array}\label{eq:w_update}
\end{equation} 

As mentioned in Section \ref{sec:MLAM}, at the inner loops, the parameters $\bm{\theta}_{\bm{X}}$ and $\bm{\theta}_{\bm{W}}$ are frozen, and are used to generate the update steps $\bm{H}_{\bm{X}}$ and $\bm{H}_{\bm{W}}$ for variables with frozen iteration numbers; therefore, the update strategy is essentially determined by the parameters $\bm{\theta}_{\bm{X}}$ and $\bm{\theta}_{\bm{W}}$. At the outer loops, we leverage the accumulated global losses to guide the parameter update for better optimization strategy through backpropagation. Let $t_{out}$ denote the update interval,  the accumulated global loss is given by 
\begin{equation}
    \mathcal{L}_{F}^s=\frac{1}{t_{out}}\sum_{t_s=(s-1)t_{out}+1}^{st_{out}}\omega_{t_s}F(\bm{W}_{t_s},\bm{X}_{t_s}),\label{eq:accumulated Global Loss}
\end{equation} 
where $\omega_{t}\in\mathbb{R}_{\geq0}$ denotes the weight associated with each outer step, and $s=1,2,\ldots,S$, with $S=T/t_{out}$ being the maximum update number for LSTM networks, and $T$ being the maximum outer steps. For every $t_{out}$ outer loop {iteration}, the accumulated global losses $\mathcal{L}_{F}^s$ is computed and is used to update $\bm{\theta}_{\bm{X}}$ and $\bm{\theta}_{\bm{W}}$ as follows
\begin{equation}
    \begin{aligned}
        \bm{\theta}_{\bm{X}}^{s+1}=\bm{\theta}_{\bm{X}}^s+\alpha_{\bm{X}}\cdot\mathrm{Adam}(\bm{\theta}_{\bm{X}}^s, \nabla_{\bm{\theta}_{\bm{X}}^s}\mathcal{L}_{F}^s),\\
        \bm{\theta}_{\bm{W}}^{s+1}=\bm{\theta}_{\bm{W}}^s+\alpha_{\bm{W}}\cdot\mathrm{Adam}(\bm{\theta}_{\bm{W}}^s, \nabla_{\bm{\theta}_{\bm{W}}^s}\mathcal{L}_{F}^s),    
    \end{aligned}
\label{update theta}
\end{equation}
where $\alpha_{\bm{X}}$ and $\alpha_{\bm{W}}$ denote the learning rates for the meta-networks $\text{{\fontfamily{cmss}\selectfont LSTM}}_{\bm{X}}$ and $\text{{\fontfamily{cmss}\selectfont LSTM}}_{\bm{W}}$, respectively. The parameters of $\text{{\fontfamily{cmss}\selectfont LSTMs}}$ are updated by the Adam method \cite{kingma2014adam}. 

\begin{rem}
 {The formulation (\ref{eq:accumulated Global Loss}) can be extended
to accommodate prior information as follows, 
\begin{equation}
\mathcal{L}^{'}=\mathcal{L}_{F}^s +\omega_{w}\left\Vert\bm{W}_{t_s}-\tilde{\bm{W}}\right\Vert _{F}^2+\omega_{x}\left\Vert\bm{X}_{t_s}-\tilde{\bm{X}}\right\Vert _{F}^2,
\end{equation}
where $\omega_{w}$ and $\omega_{x}$ denote the weights of the prior knowledge, $\left\{ \tilde{\bm{W}}\right\} $
and $\left\{ \tilde{\bm{X}}\right\} $ are the available paired training samples from historical
data.}
\end{rem}

\begin{figure}
    \centering
    \includegraphics[width=0.45\textwidth]{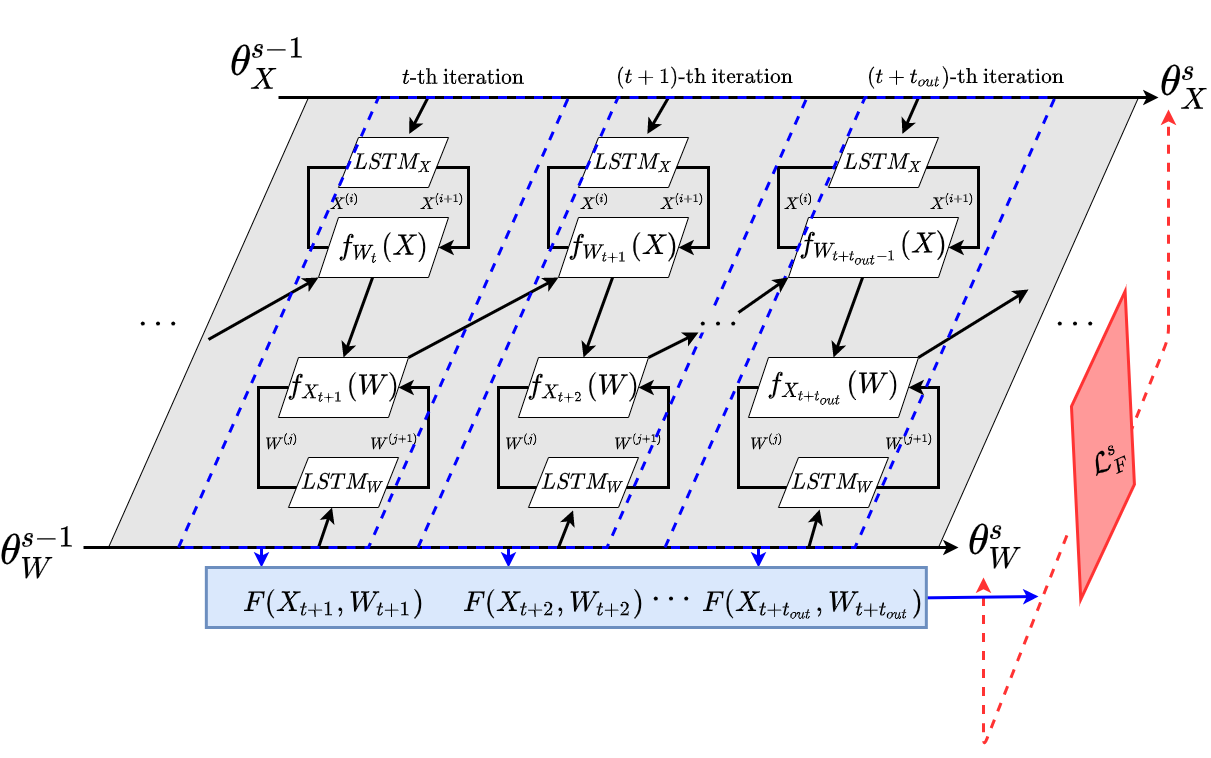}
    \caption{The outer loop iterations on sub-problems  {are} placed on the grey plane. Each iteration contains two sub-problems (dashed blue rhombus), and each sub-problem optimizes the corresponding variable at inner loop along with an LSTM  $\text{{\fontfamily{cmss}\selectfont MetaNet}}$. The global loss $F(\bm{X},\bm{W})$ is computed after each iteration. For every $t_{out}$ number of iterations, the accumulated global losses $\mathcal{L}_F^s$ (rhombus in red) is calculated and back-propagated to update the parameters $\bm{\theta}_{\bm{X}}$ and $\bm{\theta}_{\bm{W}}$ based on gradient descent manners (referred by the red dashed arrows).
    }
    \label{fig:accumulated loss}
\end{figure}

Hence, $\bm{\theta}_{\bm{X}}$ and $\bm{\theta}_{\bm{W}}$ successfully build a connection between the variable update functions and the global losses. At inner loops, $\bm{\theta}_{\bm{X}}$ and $\bm{\theta}_{\bm{W}}$ convey an extra global loss information from the outer loops for the update functions. The accumulated global losses allow the $\text{{\fontfamily{cmss}\selectfont LSTMs}}$ to be updated with respect to the mutual knowledge of  dealing with different sub-problems. Therefore, in the ground-level meta-learning, the learnt algorithm has better adaptability to the new sub-problem in the sequence. The accumulated global losses based ground-level meta-learning is depicted in Fig.\ref{fig:accumulated loss}. 

There are two merits of adopting $\mathcal{L}_F^s$ to update the parameters. First, the update with a small update interval,  {e.g.}, $t_{out}=1$, will lead to severe fluctuation. An appropriate update interval with accumulated global losses is able to effectively relieve  the factor of the outliers in training process. We shall present the related simulation results in Section \ref{sec:matrix completion}.
Second, the leveraged global losses can provide more mutual knowledge than one global loss value. The parameters are updated with the objective to minimize a partial trajectory of the global losses. Therefore, it allows the $\text{{\fontfamily{cmss}\selectfont MetaNet}}$
to learn a non-monotonic solution, where the global loss could increase at the beginning iterations but quickly decrease to better optimums on the global scope. We will discuss in details with a practical example in Section \ref{sec:GMM-demonstration}.

\begin{algorithm}[t]
    \SetAlgoLined
    
    \textbf{Input:} global loss function $F(\bm{W},\bm{X})$, local loss functions $f_{\bm{W}}(\bm{X})$ and $f_{\bm{X}}(\bm{W})$, random initialization $\bm{W}_0$, number of outer loops $T$, and number of inner loops $I$ and $J$.
    
    \textbf{Output:} Estimated variables $\bm{W}_T$, $\bm{X}_T$.
    
    % \textbf{Given:} global loss function $F(\bm{W},\bm{X})$, local loss functions $f_{\bm{W}}(\bm{X})$ and $f_{\bm{X}}(\bm{W})$:
    
    \For{$t\gets 1$, $\ldots$ T}{
    % $i, j, s = 1$
    % $s = 1$
    
    \For{$i\gets 0$, $\ldots I-1$}{
    % \While{$i\leq I$}{
        $\Delta \bm{X}=\text{{\fontfamily{cmss}\selectfont LSTM}}_{\bm{X}}(\nabla f_{\bm{W}_{t}}(\bm{X}^{(i)}),\bm{C}_{\bm{X}}^{(i)},\bm{\theta}_{\bm{X}}^s)$
        
        $\bm{X}^{(i+1)}\leftarrow \bm{X}^{(i)}+\Delta \bm{X}$\;
        % $i = i + 1$
        % }
    }
    $\bm{X}_t \gets \bm{X}^{(i)} $;
    
    Update local loss function $f_{\bm{X}_t}(\bm{W})$;

    \For{$j\gets 0$, $\ldots J-1$}{
    % \While{$j \leq J$}{
        $\Delta \bm{W} =\text{{\fontfamily{cmss}\selectfont LSTM}}_{\bm{W}}(\nabla f_{\bm{X}_{t}}(\bm{W}^{(j)}),\bm{C}_{\bm{W}}^{(j)},\bm{\theta}_{\bm{W}}^s)$
        
        $\bm{W}^{(j+1)}\leftarrow \bm{W}^{(j)}+\Delta \bm{W}$\;
        % $j = j + 1$
        % }
    }

    % \textbf{Output:} $\bm{W}_j$ as $\bm{W}_t$.
    $\bm{W}_t \gets \bm{W}^{(j)} $;
    
    Update local loss function $f_{\bm{W}_t}(\bm{X})$;
    
    Update global loss function $F(\bm{W}_t,\bm{X}_t)$;
    
    \For{$s\gets 1$, $\ldots t/t_{out}$}{
    % \While{$s\le t/t_{out}$}{
    $\mathcal{L}_{F}^s=\frac{1}{t_{out}}\sum_{t_s=(s-1)t_{out}+1}^{st_{out}}\omega_{t_s}F(\bm{W}_{t_s},\bm{X}_{t_s})$
    
    $\bm{\theta}_{\bm{X}}^{s+1}=\bm{\theta}_{\bm{X}}^s-\alpha_{\bm{X}}\nabla_{\bm{\theta}_{\bm{X}}^s}\mathcal{L}_{F}^s$
    
    $\bm{\theta}_{\bm{W}}^{s+1}=\bm{\theta}_{\bm{W}}^s-\alpha_{\bm{W}}\nabla_{\bm{\theta}_{\bm{W}}^s}\mathcal{L}_{F}^s$
    
    % $s=s+1$
    }
    }
\caption{\label{alg:MLAMOpt}General Structure of MLAM algorithm for solving one problem
}
\end{algorithm}

At this stage, the proposed MLAM algorithm is summarized in Algorithm \ref{alg:MLAMOpt}. Specifically, the algorithm starts from a random initialization $\bm{W}_0$ at the beginning of an outer loop. At the $t$-th outer loop, it contains two inner loops for updating $\bm{X}$ and $\bm{W}$, respectively. Each inner loop starts from a random initialization $\bm{X}^{(0)}$ and $\bm{W}^{(0)}$ and updates variables based on equations (\ref{eq:x_update}) and (\ref{eq:w_update}) and then repeats for $I$ and $J$ times, respectively. In this paper, we set $I=J=t_{in}$ in which $t_{in}$ indicates the maximum iteration number at the inner loops. The choices for $t_{out}$ and $t_{in}$ will be discussed in Section \ref{sec:matrix completion}. In practice, it is reasonable to set different values for $I$ and $J$ according to the demand of the objective in different problems. At the end of each inner loop, the output of this inner loop is regarded as $\bm{X}_t$ or $\bm{W}_t$ at the $t$-th outer loop, and is then assigned to generate sub-problem $f_{\bm{X}_t}$ and $f_{\bm{W}_t}$, respectively. For every $t_{out}$  {step} at outer loops, the parameters $\bm{\theta}_{\bm{X}}$ and $\bm{\theta}_{\bm{W}}$ are updated following equations (\ref{update theta}). Finally, our MLAM algorithm will stop when $t$ reaches the maximum outer loop number $T$. 

% \begin{rem}
% In the rest of this paper, we set $I=J=t_{in}$ for demonstration convenience. The choices for $t_{out}$ and $t_{in}$ are discussed in Section \ref{sec:parameter tunning}. In practice, it is reasonable to set different values for $I$ and $J$ according to the demand of the objective in different problems.
% \end{rem}

Comparing to the deep unfolding algorithms, the fundamental  {differences} of MLAM method are mainly three-fold: 1) replacing the variable update function within each iteration by a meta-network instead of mapping the whole procedure at each iteration by black-box based network operator, thus the inner loop at each iteration is interpretable; 2) focusing on learning a new strategy for the whole iterations instead of following the basics of the original strategy with end-to-end network behavior on trainable parameters, leading to a better explainability on the optimization manner; and 3) is feasible for unsupervised learning while the deep unfolding methods are mostly supervised learning. Therefore, the proposed MLAM method highlights advances on the model explainability and the improvements of combining the advantages of learning-based and model-based approaches. We believe this is a further step ahead that makes a higher level of learning than the deep unfolding, where the learning objective is no longer the trainable parameters but the whole optimization strategy.  

In a summary, a hierarchical LSTM-based MLAM model is proposed in this section. It contains two (or more for multiple variables if needed) LSTM networks which perform optimization at inner loops with frozen parameters; their parameters are updated during outer loops with respect to minimizing accumulated global losses. Therefore, the original structure of algorithms well-established in the field is maintained while the performance is improved by the meta-learned algorithmic rule. 

\section{Applications in Typical Problems}
% In this section, we show two exemplar applications of our MLAM method on solving bi-linear inverse problem and non-linear problem in non-convex set. More specifically, we apply our proposed MLAM method on the matrix completion problem and on the Gaussian mixture model (GMM) problem. In this paper, the goal of these two application is a proof of the concept that the MLAM is effective and feasible on solving non-convex problems, instead of emphasising the practical implementation.

% The analysis of feasibility of applying our model to solve these problems is given. 
% \textcolor{red}{We further show that how our learned algorithm iterates non-optimums, and perform better numerical results than iterating optimums in demo case.}
%\subsubsection{Example 1 (Dictionary Learning)}

In this section, we shall apply the proposed MLAM approach to a bi-linear inverse problem and a non-linear problem. For those problems, the model-based methods perform less effective and learning-based methods typically do not work due to the lack of sufficient labelled data. Specifically, matrix completion and Gaussian mixture model (GMM) problems are analyzed in this paper as two representatives.

\subsection{Bi-linear Inverse Problem: Matrix Completion}\label{sec:MF}
% One of the most well known optimization problems over the intersection of two variable sets is the bi-linear
% inverse problem, which is a challenging and representative task of
% optimizing non-convex problem, with a variety of important practical examples
% including blind deconvolution \cite{hopgood2003blind}, blind source
% separation \cite{o2005survey}, dictionary learning \cite{yu2019bilinear},
% and matrix completion \cite{donoho2004does}. 
The bi-linear inverse problem is a typical optimization problem whose variables are within an intersection of two sets. Many non-convex optimization problems can be cast as bi-linear inverse problems, including low-rank matrix recovery \cite{jain2013low,gross2011recovering,wen2012solving,hardt2014understanding,koren2009matrix}, dictionary learning \cite{tosic2011dictionary,aharon2006k,yu2019bilinear}, and blind deconvolution \cite{zhang2017global,stockham1975blind,hopgood2003blind}. 
% Typically, a bi-linear inverse problem aims to estimate the variable pair $(\bm{W},\bm{X})$ from some observed samples $\bm{Y}$ with a given bi-linear mapping $\bm{F}(\bm{w},\bm{x})=\bm{y}$.
For example, in  bind deconvolution and in the matrix completion, ${F}(\cdot, \cdot)$ in \eqref{overall problem} represents circular convolution and matrix product, respectively. 
% Most of these problems enjoy good landscapes where their second-order stationary points are all global optimums \cite{jain2013low,jain2017non}. The benign landscapes can be typically guaranteed by assuming certain constraints on variable sets, such as sparse constraint \cite{dai2009subspace,yang2015sparse,fan2017sparse}, semi-defined constraint \cite{fan2017sparse,hardt2014understanding}, and rank limitation \cite{jain2013low,jain2017non}.

Next we focus on the matrix completion \cite{koren2009matrix,lu2012robust,wen2012solving,candes2009exact,hardt2014understanding} as a representative bi-linear inverse problem to demonstrate how to apply MLAM to solve this type of problems. Matrix completion is a class of tasks that aims to recover missing entries in a data matrix
\cite{candes2009exact,lu2012robust,fan2017sparse} and has been widely applied in practice including recommending system \cite{koren2009matrix} and collaborative filtering \cite{zhou2008large}. Typically, it is formulated as a low-rank matrix recovery problem in which the matrix to be completed is assumed to be low-rank with given rank information. 

As for optimization, the low-rank matrix completion problem is usually formulated as the multiplication of two matrices and then converted into two corresponding sub-problems which are generally strongly convex \cite{choudhary2013identifiability,gross2011recovering,jain2013low}. Then, gradient descent based methods, such as alternating
least square (ALS) \cite{kroonenberg1980principal} and stochastic
gradient descent (SGD) \cite{gemulla2011large}, are often applied on solving each sub-problem and can achieve good performance under certain conditions. 
However, satisfactory results are not universally guaranteed.  
On the one hand, the performance depends on a set of factors, including initialization strategies, the parameter setting of gradient descent algorithms, the sparsity level of the low-rank matrix, etc. On the other hand, the assumptions behind these constraints are stringent, such as the feature bias on variables for sparse subspace clustering mechanism \cite{yang2015sparse,fan2017sparse}, which are not common in practice. 

Therefore, the bi-linear inverse problem is basically ill-posed and thus has always been a challenging task. Consequently, the performance of the traditional model-based methods is highly limited by the prior knowledge on models and structural constraints. Meanwhile, the strong non-convexity and constraints also limit the application of the deep learning and unfolding techniques in this type of problems. 

In this section, we further consider several more realistic and therefore challenging scenarios, including high-rank matrix completion, matrix completion without given rank knowledge, and mixed rank matrix completion problems, where the existing model-based and even learning-based approaches fail.

The matrix completion problem is then formulated as \cite{Rennie2005Fast},
\begin{equation}
    \min_{\bm{U},\bm{V}} F(\bm{U},\bm{V}):=\frac{1}{2}\left\Vert \mathcal{P}_{\Omega}(\bm{R}-\bm{U}\bm{V}^{T})\right\Vert _{F}^{2}+\frac{\lambda}{2}(\left\Vert \bm{U}\right\Vert _{F}^{2}+\left\Vert \bm{V}\right\Vert _{F}^{2}),
\label{eq:MF}
\end{equation}
where the projection $\mathcal{P}_{\Omega}(\cdot)$ preserves the observed elements defined by $\Omega$ and replaces the missing entries with $0$, and $\lambda$ is the weight parameter of the regularizers. The matrix completion problem is typically formulated as a low-rank matrix recovery problem, which parametarizes a low-rank matrix $\bm{R}\in\mathbb{R}^{z\times q}$ as a multiplication of two matrices $\bm{U}\bm{V}^T$ with $\bm{U}\in\mathbb{R}^{z\times p}$, $\bm{V}\in\mathbb{R}^{q\times p}$ and $p\leq\min(z,q)$.
\begin{rem}\label{rank remark}
When taking the matrix multiplication $\bm{R}=\bm{U}\bm{V}^T$, the parameter $p$ is set to the rank of $\bm{R}$, which is typically known. If the rank is not provided, the problem will be much more difficult and the existing methods would be unworkable. In Section \ref{sec:matrix completion}, we will verify that the proposed MLAM still works properly when $p$ is unknown.
\end{rem}

Here we define (\ref{eq:MF}) as the overall problem, and $F(\bm{U},\bm{V})$ as the global loss function. It is obvious that the overall problem is not convex in terms of $\bm{U}$ and $\bm{V}$, but the sub-problems are convex when fixing one variable and updating the other. Therefore, we split (\ref{eq:MF}) into two sub-problems in quadratic form, fixing one variable in (\ref{eq:MF}) and updating the other one, referring to $f_{\bm{U}}(\bm{V})$ and $f_{\bm{V}}(\bm{U})$. The ALS and SGD methods can be applied  by iteratively minimizing the two sub-problems.  

Meanwhile, according to Algorithm \ref{alg:MLAMOpt}, our MLAM method can be directly applied to solve the problem (\ref{eq:MF}) using two LSTM networks $\text{{\fontfamily{cmss}\selectfont LSTM}}_{\bm{U}}$ and $\text{{\fontfamily{cmss}\selectfont LSTM}}_{\bm{V}}$ with parameters $\bm{\theta}_{\bm{U}}$ and $\bm{\theta}_{\bm{V}}$, to optimize matrices $\bm{U}$ and $\bm{V}$, respectively. The variable update equations are given by
\begin{equation}
\begin{array}{c}
\bm{U}^{(i)}=\bm{U}^{(i-1)}+\bm{H}_{\bm{U}}^{(i-1)},\\
\bm{V}^{(j)}=\bm{V}^{(j-1)}+\bm{H}_{\bm{V}}^{(j-1)},      
\end{array}\label{eq:UV_update}
\end{equation} 
where $\bm{H}_{\bm{U}}$ and $\bm{H}_{\bm{V}}$ are the outputs of $\text{{\fontfamily{cmss}\selectfont LSTM}}_{\bm{U}}$ and $\text{{\fontfamily{cmss}\selectfont LSTM}}_{\bm{V}}$, respectively.
Two $\text{{\fontfamily{cmss}\selectfont LSTM}}$ networks are updated for every $t_{out}$ outer loop steps, via back-propagating accumulated global losses $\mathcal{L}^{s}_{F}$ according to equation (\ref{eq:accumulated Global Loss}). In this way the updating rule of the learned algorithm could be adjusted to find gradient descent steps $\bm{H}_{\bm{U}}$ and $\bm{H}_{\bm{V}}$ for minimizing $F(\bm{U},\bm{V})$ adaptively.

% In Fig \ref{???}, we present ?? examples of loss trajectories on loss surface for matrix completion problem that has high non-convexity. It is clear that our MLAM method successfully converge to global optimum in all examples, meanwhile, AM methods are trapped in local optimum. 
The advantages of our MLAM model for the matrix completion problem may be explained by the replacement of updating step function $\phi$ by the neural networks. In the AM methods, when a local optimum $(\bm{U}^{'},\bm{V}^{'})$ is reached, the gradient of the variables equals to zero, \textit{e.g.,} $\frac{\partial f_{\bm{V}^{'}}(\bm{U})}{\partial\bm{U}}|_{\bm{U}{=\bm{U}'}}=\bm{0}$. At this stage, the update function, which is determined by the gradients of variables, will be stuck at local optimum. While in our MLAM model, our update functions are determined by their parameters, $\bm{\theta}_{\bm{U}}$ and $\bm{\theta}_{\bm{V}}$, which are further determined by leveraging on partial global loss trajectories across the outer loop steps. This major difference mainly brings two benefits to our MLAM method. One advantage is that even at local optimum points, our MLAM can still provide a certain step update on variables. This can be understood that even when one of the input $\frac{\partial f_{\bm{V}^{'}}(\bm{U})}{\partial\bm{U}}|_{\bm{U}{=\bm{U}'}}=\bm{0}$, $\text{{\fontfamily{cmss}\selectfont LSTM}}_{\bm{U}}$ still obtains some non-zero outputs given a non-linear function of zero input, cell state, and parameters. Another advantage is that the leveraged global loss leads to a smooth optimization on a global loss landscape, which allows the learned algorithm to be essentially guided by the inductive bias from a smoother transform of the global loss landscape. In Section V, besides traditional low rank matrix completion problem, we further evaluate MLAM in matrix completion problems in the case of high rank, unknown rank and mixed rank ( {a set of matrices} with different ranks, from low rank to full rank).   

\subsection{Non-linear Problem: Gaussian Mixture Model}\label{sec:GMM-demonstration}
% In contrast to the bi-linear inverse problem, the non-linear problem is more challenging due to the fact that there is no analytic geometry for this type of problems and sub-problems. GMM is one of the most classical and significant problems. It is a probabilistic model for representing observed data samples as weighted Gaussian components \cite{reynolds2009gaussian}. %Each Gaussian is recognized as one cluster in the mixture model, which is parameterized by three variables. 
Different from the bi-linear inverse problems, optimization over the intersection of two variables that has a non-linear mapping between variables and observed samples also plays a significant role in statistical machine learning, including Bayesian model \cite{Li2015Multispectral}, graphic model \cite{Chandrasekaran2011Rejoinder,stankovic2020graph}, and finite mixture model \cite{dempster1977maximum}. The sub-problem in non-linear problem typically has no closed-form solution and the overall problem possesses many local  {optimums}. Therefore, it is difficult to guarantee the convergence of the model-based approaches, especially for the global optima, as well as intractable to obtain labelled data for the existing learning based methods. 

GMM \cite{xu1996convergence,hosseini2015matrix,reynolds2009gaussian} is one of the most important probabilistic models in machine learning. GMM problem, consisting of a set of Gaussian distributions in the form of weighted Gaussian density components, is usually estimated by maximum log-likelihood method \cite{reynolds2009gaussian}. The variables in GMM problem possess a non-linear mapping to the observations. Many methods have been proposed to solve the GMM problem, i.e., maximising the likelihood of GMM problems, such as conjugate gradients, quasi-Newton and Newton \cite{redner1984mixture}. However, these methods typically perform inferior to the one called expectation-maximization (EM) algorithm \cite{dempster1977maximum,xu1996convergence}. One possible reason is due to the non-convexity and non-linearity of the GMM problem that requires a sophisticated step descent strategy to find a good stationary point. On the other hand, EM algorithm omits the hyperparameter related to the step size by converting the origin estimation problem of maximising likelihood into a relaxed problem where a lower bound is maximized monotonically and analytically. Even though the EM algorithm has been widely applied in the GMM problem, it also suffers from the aforementioned non-convexity and non-linearity. When the non-convexity is high (referring to some real-world scenarios: the number of observation is not sufficient, dealing with high-dimensional data \cite{ge2015learning} and large number of clusters), the convergence is not guaranteed and the performance significantly degrades. Many works attempt to replace EM algorithm through reformulating GMM as adopting matrix manifold optimization \cite{hosseini2015matrix, li2019general}, and also learning based method in high dimensions \cite{ge2015learning}.

Detailed descriptions of GMM problems can be found in \cite{mclachlan2004finite}. Given a set of $G$ i.i.d samples $\bm{X}=\{\bm{x}_g\}_{g=1}^G$, each entry $\bm{x}_g$ is a $D$-dimensional data vector. Then, a typical optimization when using the GMM to model the samples is to maximize the log-likelihood (MLL) \cite{reynolds2009gaussian}, which is equivalent to minimize the Kullback–Leibler divergence from the empirical distribution. The parameters of the GMM can then be optimized as follows,
\begin{equation}
     \max_{\{{\pi}_k,\bm{\mu}_k,\bm{\Sigma}_k\}_{k=1}^{K}}\log{p(\bm{X})}=\sum_{g=1}^{G}\log{\sum_{k=1}^K}\pi_k\mathcal{N}(\bm{x}_g\arrowvert\bm{\mu}_{k},\bm{\Sigma}_{k})\label{Log-max-likelihood},
\end{equation}
% \begin{equation}\label{eq:p-gaussian}
%  p(\bm{X})=\prod_{n=1}^{N}\sum_{k=1}^{K}\pi_k \mathcal{N}(\bm{x}_n\arrowvert\bm{\mu}_{k},\bm{\Sigma}_{k}),  
% \end{equation} 
where
\begin{equation}
\mathcal{N}(\bm{x}_g\arrowvert\bm{\mu}_{k},\bm{\Sigma}_{k})=\frac{\exp \{-\frac{1}{2}(\bm{x}_g-\bm{\mu}_k)^T\bm{\Sigma}_k^{-1}(\bm{x}_g-\bm{\mu}_k)\}}{(2\pi)^{D/2}\arrowvert \bm{\Sigma}_k \arrowvert^{1/2}}.  
\end{equation}
In \eqref{Log-max-likelihood}, for the $k$-th Gaussian component, $\bm{\mu}_k$ is the mean vector and defines the cluster centre, covariance $\bm{\Sigma}_k$ denotes the cluster scatter, $\pi_k$ represents mixing proportion with $\sum_{k=1}^{K}\pi_k=1$, and $\arrowvert \bm{\Sigma}_k \arrowvert$ represents the determi$>$1t of $\bm{\Sigma}_k$.

However, it is intractable to directly obtain a closed-form solution that maximizes $\log{p(\bm{X})}$ in (\ref{Log-max-likelihood}). The key difficulty is that by differentiating $\log{p(\bm{X})}$ (summation of logarithmic summation) and equalizing it to 0, each parameter is intertwined with each other. Gradient descent methods in an AM manner can alternatively solve \eqref{Log-max-likelihood}, but they typically perform inferior to the EM algorithm \cite{dempster1977maximum,xu1996convergence}. 

We should also point out that the EM algorithm still updates the GMM parameters in an AM manner, i.e., iterating to optimize over each parameter at which its gradient equals to $0$ with the other parameters being frozen. Furthermore, the EM algorithm, together with other first-order methods, has been proved to converge to arbitrary bad local optimum almost surely \cite{jin2016local}. As we have discussed before, this AM strategy can be improved by replacing the frozen updating rule that searches optimum in a local landscape, by a less-greedy rule that updates variables up to global scope knowledge on the loss landscape of the global objective function.

Therefore, we propose to solve GMMs by adopting our MLAM method, which directly applies a learning-based gradient descent algorithm to the original ML problem (\ref{Log-max-likelihood}) without any extra constraints. In this scenario, we consider the GMM problem with covariance $\bm{\Sigma}$ being given; hence the MLL estimation of GMM is presented in terms of negative log-likelihood as follows
\begin{equation}
  \min_{\{{\pi}_k,\bm{\mu}_k\}_{k=1}^K}F_{ne}(\bm{\pi},\bm{\mu})=-\sum_{g=1}^{G}\log{\sum_{k=1}^K}\pi_k\mathcal{N}(\bm{x}_g\arrowvert\bm{\mu}_{k}, \bm{\Sigma}_{k}).\label{Ne-ML-GMM}  
\end{equation} 

In this case, we treat the (\ref{Ne-ML-GMM}) as our overall problem, and split it into two sub-problems $f_{\bm{\pi}}(\bm{\mu})$ and $f_{\bm{\mu}}(\bm{\pi})$.  Define vector $\bm{\pi}\in\mathbb{R}^{K}$ and matrix $\bm{\mu}\in\mathbb{R}^{K\times D}$, where $K$ is the maximum number of clusters and $D$ is the dimensionality of samples. In our MLAM framework, we build two \text{{\fontfamily{cmss}\selectfont MetaNets}} $\text{{\fontfamily{cmss}\selectfont LSTM}}_{\bm{\pi}}$ and $\text{{\fontfamily{cmss}\selectfont LSTM}}_{\bm{\mu}}$ with parameters $\bm{\theta}_{\bm{\pi}}$ and $\bm{\theta}_{\bm{\mu}}$  to update $\bm{\pi}$ and $\bm{\mu}$ as follows
\begin{equation}
\begin{array}{c}
\bm{\pi}^{(i)}=\bm{\pi}^{(i-1)}+\bm{H}_{\bm{\pi}}^{(i-1)},\\
\bm{\mu}^{(j)}=\bm{\mu}^{(j-1)}+\bm{H}_{\bm{\mu}}^{(j-1)},      
\end{array}\label{eq:pimu_update}
\end{equation}
where $\bm{H}_{\bm{\pi}}$ and $\bm{H}_{\bm{\mu}}$ are the outputs of the two LSTM neural networks respectively. 

Similar to applying MLAM in matrix completion, we start from a random initialization $\bm{\mu}^0$ and $\bm{\pi}^0$. During our MLAM procedure given in Algorithm \ref{alg:MLAMOpt}, $\bm{\pi}_k$ and $\bm{\mu}_k$ are updated based on equation (\ref{eq:pimu_update}) for $t_{in}$ steps in inner loops respectively. Meanwhile, we back-forward the accumulated global losses with respect to the negative log-likelihood of GMM, referring to $\mathcal{L}^{s}_{F}$, for every $t_{out}$ steps on the outer loops. In this way, the algorithm is updated based on the global scope knowledge about the global losses across outer loop iterations.  

\begin{figure}[t]
    \centering
     \includegraphics[width=0.45\textwidth]{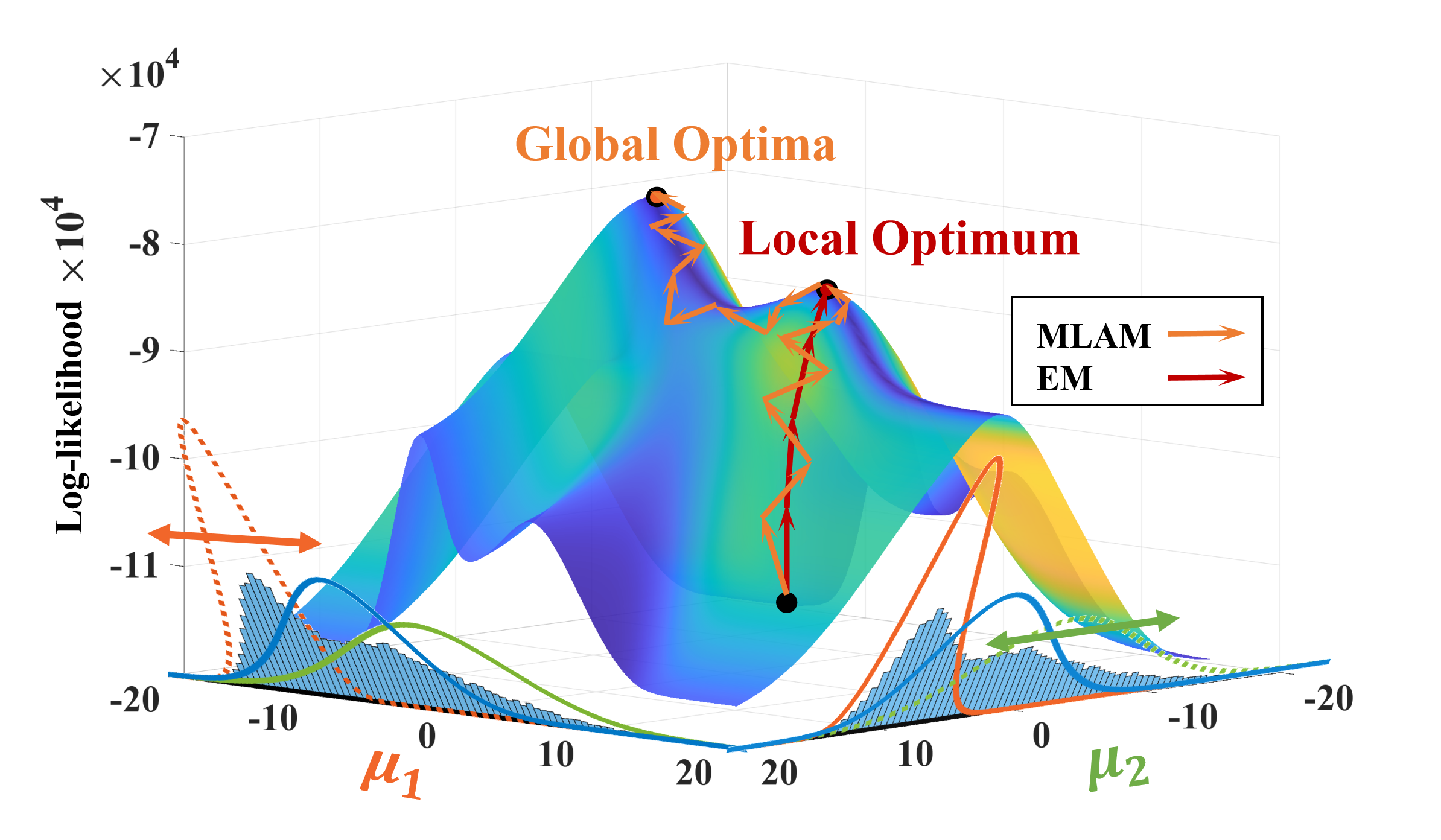}
    \caption{An illustration of converge trajectories achieved by the proposed MLAM and EM methods in a simple GMM problem. In this GMM, the samples were obtained by three Gaussian distributions of parameters $\{\mu_1=-3, \mu_2=3, \mu_3=0\}$, $\{\sigma_1=1, \sigma_2=10, \sigma_3=5\}$, and $\{\pi_1=\pi_2=\pi_3=\frac{1}{3}\}$. For the convenience of visualization, we here show the surface of log-likelihood and converge trajectories with regard to $\mu_1$ and $\mu_2$, whereas setting the other parameters to the ground-truth when optimizing GMMs; this means the global optima exists at $\mu_1=-3$ and $\mu_2=3$.}
    \label{fig:GMMIllu}
\end{figure}

Considering that the existing numerical gradient-based solutions typically perform less effectively and accurately than the EM algorithm \cite{xu1996convergence}, we focus on comparing our MLAM method and the EM algorithm. EM algorithm converts maximization of the log-likelihood into maximization on its lower bound; hence it has closed-form formulations to implement an AM strategy on its maximization step for variable updating. Nevertheless, our MLAM method directly optimizes the cost function, i.e., the log-likelihood, and the variables are updated constantly up to the global scope knowledge extracted by $\text{{\fontfamily{cmss}\selectfont LSTMs}}$. 
In Fig.\ref{fig:GMMIllu} we show the convergence trajectory of our MLAM and EM algorithm on the geometry of a GMM problem with three clusters. As mentioned in Section \ref{LSTM MLAM}, the proposed MLAM with accumulated global losses allows the learnt algorithm  {converges} in a non-monotonic way. It can be seen that the EM algorithm (red arrows) quickly converges to the local optimum and stops at it. In contrast, the MLAM approach (orange arrows), first going to the local optimum yet, is able to escape from the local optimum and converges to the global optima. Specifically, when encountering local optimum, the proposed MLAM first goes down on the geometry, and then moves towards to the global optima, thus escaping from the local optimum. This reveals that the learnt algorithm does not request each step moving towards to the most ascending direction, but the global losses on a partial trajectory should be minimized, recalling the accumulated global losses minimization $\mathcal{L}_F^s$ in equation (\ref{eq:accumulated Global Loss}).  Therefore, the MLAM enjoys significant freedom to learn a non-monotonic algorithm for convergence in terms of the update strategy on $\mathcal{L}_F^s$. We also shall point out that this also results in the fluctuations on the trajectory, which could increase the needed iterations when the geometry is smooth and benign.   

\section{Experiments}
In this section, we will present simulation results on the aforementioned matrix completion and GMM problems to validate the effectiveness and efficiency of our MLAM method.

For the experimental settings, our \text{{\fontfamily{cmss}\selectfont MetaNets}} employ two-layer $\text{{\fontfamily{cmss}\selectfont LSTM}}$ networks with $500$ hidden units in each layer. Each network is trained by minimizing the accumulated loss functions according to equation (\ref{eq:accumulated Global Loss}) via truncating backpropagation through time (BPTT) \cite{werbos1990backpropagation}, which is a typical training algorithm to update weights in RNNs including LSTMs. The weights of $\text{{\fontfamily{cmss}\selectfont LSTM}}$ are updated by Adam \cite{kingma2014adam}, and the learning rate is set to $10^{-3}$. In all simulations, we set $\omega_{t_s}=1$ for simplicity. The parameters of LSTM networks are randomly initialized and continuously updated through the whole training process. For evaluation, we fix the parameters of our MLAM model and evaluate the performance on the testing datasets.

\subsection{Numerical Results on Matrix Completion}\label{sec:matrix completion}
In this subsection, we consider learning to optimize synthetic $D$-dimensional matrix completion problems. We take $D=10$ and $D=100$ to evaluate algorithms for both small and large scale cases. For each matrix completion problem, the ground truth matrix $\bm{R}\in \mathbb{R}^{D\times D}$ is randomly and synthetically generated with rank p. Meanwhile, the observation $\bm{R}_{S}=\mathcal{P}_{\Omega}(\bm{R})$ is generated by randomly setting a certain percentage of entries in $\bm{R}$ to be zeros, and the non-zero fraction of entries is the observation rate. Matrix completion for $\bm{R}$ is then achieved by solving the low-rank matrix recovery on $\bm{R}_{S}$ in the form of equation (\ref{eq:MF}) in Section \ref{sec:MF}. The two factorized low-dimensional matrices $\bm{U}\in\mathbb{R}^{D\times p}$ and $\bm{V}^{D\times p}$ are then used to generate reconstruction of the ground truth matrix, denoted by $\hat{\bm{R}}=\bm{U}\bm{V}^{T}$. The evaluation criterion is given by Relative Mean Square Error (RMSE) of 
\begin{equation*}
    \textrm{RMSE}=\frac{\Vert \bm{R}-\hat{\bm{R}} \Vert_F}{\Vert \bm{R}\Vert_F}.\label{eq:rmse}
\end{equation*}

 {As aforementioned, the classical AM-based ALS \cite{kroonenberg1980principal} and SGD  \cite{gemulla2011large} approaches, as well as the learning-based deep matrix factorization (DMF) \cite{fan2018matrix} and unfolding matrix factorization (UMF) methods \cite{mai2021ghost}, have been adopted for comparisons.  The parameters of all compared methods are carefully adjusted to present their best performances.}

Different simulation scenarios on matrix completion are evaluated comprehensively. The detailed simulation settings include: i) each simulation contains a set of $200$ matrix completion problems, half of which are employed as training samples for parameter update on $\text{{\fontfamily{cmss}\selectfont LSTM}}$, while the remaining $100$ matrix completion problems are used to evaluate the performance as testing samples; ii) we set $T=100$ as total alternating steps for each problem; iii) the averaged RMSE over 100 testing samples is used for evaluation; iv) in all the training and testing processes, the ground truth matrix $\bm{R}$ is not given, and is only used to evaluate performance after the optimizing processes.

\subsubsection{Parameter setting}\label{sec:parameter tunning}
The numbers of variable update steps on inner loops $t_{in}$ and update interval $t_{out}$ are the two most important hyper-parameters. Different settings on $t_{in}$ and $t_{out}$ are thus tested at first to provide a brief guidance on the choices of $t_{in}$ and $t_{out}$. 
 
\begin{figure}[t]
    \centering
     \includegraphics[width=0.3\textwidth]{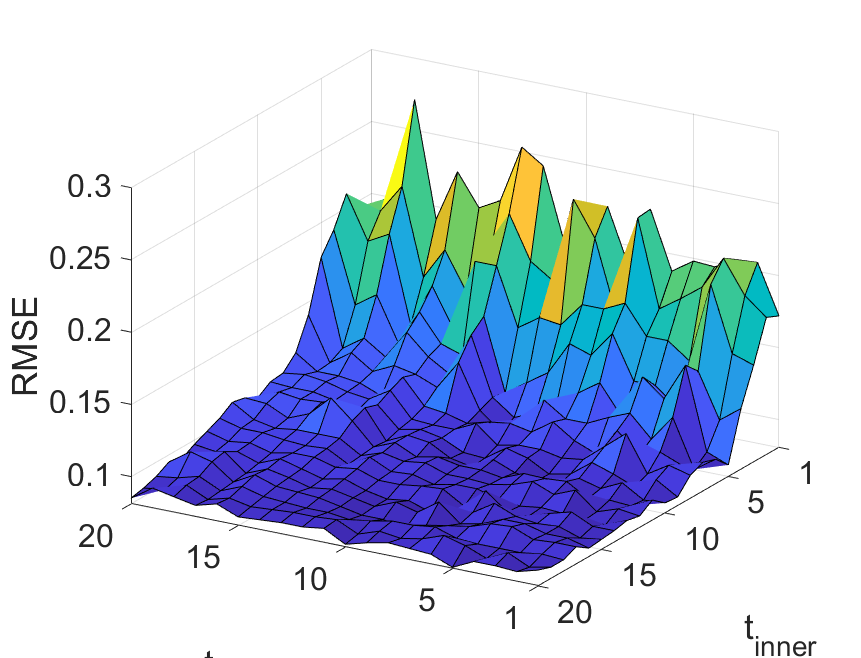}
    \caption{Performance of the proposed MLAM method on rank-5 matrix completion problem with different parameter combinations of $t_{in}$ and $t_{out}$ which range from 1 to 20 with step size 1.}
    \label{fig:MF hyper}
\end{figure}

Empirically, there is a trade-off between performance and efficiency. Here we set $t_{in}$ and $t_{out}$ to both vary from 1 to 20 with $20\%$ observation rate, and there are therefore $20\times20=400$ different parameter combinations to be evaluated for rank-5 matrix completion problems. The performance of these $400$ simulations are shown in Fig. \ref{fig:MF hyper}. We can see that for all choices of $t_{out}$, the increase on $t_{in}$ leads to significant improvements on performance when $t_{in}\leq10$, however, further increasing on $t_{in}$ witnesses little gain on performances. At the same time, the variations of $t_{out}$ have less impact on RMSE results, while larger choices ($t_{out}\geq10$) bring improved stability (there are less fluctuations when $t_{in}\geq10$, and $t_{out}\geq10$). Fig. \ref{fig:MF hyper} indicates that a sufficient number of update steps for inner and outer loops play  {a} significant role in these optimization processes. We can see that when either $t_{in}$ or $t_{out}$ is small (less than 5), the optimization does not perform well enough.
A larger choice of $t_{in}$, however, typically brings higher computational cost. Thus in the rest of this paper, we set $t_{in}=10$ and $t_{out}=10$ as the default parameter setting.

It is understandable that $t_{in}$ directly determines the number of variable update steps within each inner loops. When $t_{in}$ is small, the learned updating rule needs to optimize variables in a few steps, however this could be intractable in general. 
Meanwhile, the accumulated global losses depend on $t_{out}$ at outer loops which indicates the length of the trajectory of global losses at outer loops. Therefore, a large enough $t_{out}$ could provide sufficient global losses trajectory for parameter update.
According to the two-levels meta-learning in our MLAM model, sufficient update steps at inner and outer loops can ensure that each level of meta-learning works well. We infer that small $t_{in}$ may limit the ground-level meta-learning corresponding to inner loops, making it unable to extract effective sub-problems across knowledge with merely few update steps, and small $t_{out}$ could cause that the upper-level meta-learning corresponding to outer loops becomes less stable due to the lack of updates on parameters.

\subsubsection{Standard matrix completion}
In this part, we will first evaluate all methods on low-rank matrix completion tasks and then apply them on high rank matrix completion tasks.  {In these simulations the rank of the matrices is given. In this subsection, the main goal is the proof of the concept of the proposed MLAM. We evaluate four existing approaches, including conventional model-based methods and the state-of-the-art learning-based methods for comparison.} 

In Table \ref{tab:MF rank-5}, average RMSE of the five evaluated methods on four sets of 100-dimensional rank-5 matrix completion problems have been reported. Different sets have different observation rates, including $20\%$, $40\%$, $60\%$ and $80\%$.  {From Table \ref{tab:MF rank-5}, 
it is clear that our MLAM algorithm significantly outperforms all the existing methods, especially on high observation rate scenarios. It is also noticeable that when the observation rate is $20\%$, both of the model-based methods (ALS and SGD) and learning-based methods (DMF and UMF) work not well, while our MLAM method achieves good reconstruction with RMSE $<0.1$.} 

\begin{figure}[htbp]
  \centering
  \includegraphics[width=0.5\linewidth]{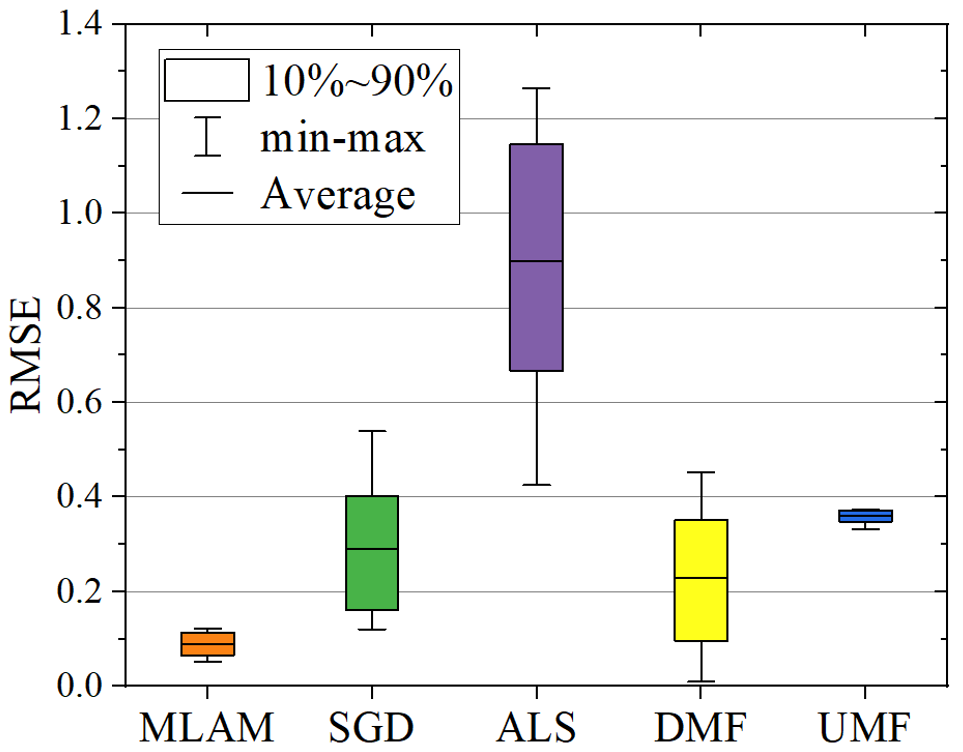}\\
  \caption{The RMSE variance of the evaluated methods on rank-5 matrix completion tasks with $20\%$ observation rate.}\label{MLAM_Variance}
\end{figure}

 {In Fig.\ref{MLAM_Variance}, we present the RMSE variance of the five tested methods on 100 trails of rank-5 matrix completion tasks with $20\%$ observation rate. It can be observed that the MLAM obtains the best performance while keep the variance relatively small. It is noticeably that the ALS approach fluctuates severely, while the SGD and DMF approaches present relatively large variance comparing to  the MLAM. Though the UMF method gains robust results with small variance, the accuracy is significantly lager than the MLAM.}

From these simulation results, we can conclude that in the classic low-rank matrix completion problem, our MLAM method has shown better performances than the comparison methods, especially when the observation rate is low.

\begin{table}[t]
\centering
\caption{ {RMSE of rank-5 matrix completion with different observation rates $\Theta$.}}
\label{tab:MF rank-5}
    \begin{tabular}{|c|c|c|c|c|}
      \hline
      \diagbox[width=\dimexpr \textwidth/8+4\tabcolsep\relax, height=0.55cm]{ Methods }{$\Theta$}
                   & 0.2   & 0.4   & 0.6   & 0.8     \\ \hline
MLAM              & \textbf{0.089} & \textbf{0.057} & \textbf{0.045} & \textbf{0.003}  \\ \hline
SGD               & 0.292  & 0.213  & 0.136  & 0.072   \\ \hline
ALS               & 0.864  & 0.717  & 0.423  & 0.228   \\ \hline
 {DMF}              & 0.228  & 0.091  & 0.057  & 0.025   \\ \hline
 {UMF}              & 0.363  & 0.359  & 0.335  & 0.312   \\ \hline

\end{tabular}
\end{table}

% \subsubsection{Scenario 3: High rank matrix completion}
We then consider more challenging matrix completion problems. Generally, the matrix completion problem is assumed to be solved as a low-rank matrix completion problem. Here we are aiming to solving high-rank or even full-rank matrix completion problems without adding any further assumptions. In this case, we test all the five approaches on six set of 100-dimensional matrix completion problems with $20\%$ observation rate, whose ranks are ranged from 5 to 100. 

The averaged RMSE results of matrix completion problems with variant rank are listed in Table \ref{tab:MF high-rank}.  {There is a clear trend of decreasing performances of the comparison methods when the rank of matrices increases. Noticeably, ALS method is no longer workable when the rank is larger than 10. SGD, DMF and UMF  methods fail after the rank reaches 40. This can be understood that these approaches are typically assumed specific penalty terms recalling to the low-rank property. However, the proposed MLAM method shows different results: the higher the rank, the smaller the RMSEs. We can conclude that our algorithm is capable of solving the matrix completion problem in high-rank, even full-rank scenarios, without any extra constraints, while standard methods typically fail.}

The execution time of the evaluated methods is compared
in Table \ref{tab:time}. It can be seen that the
proposed MLAM approach has larger time cost than the compared methods, but the difference is within second-level. This is caused by the alternatively utilizing network at each iteration. There is a trade off between the time consuming and performance: the higher the accuracy, the larger the time consumption. In this paper, the main goal is the proof of the concept that the MLAM approach could provide better performance on solving multi-variable non-convex optimization problem.
\begin{table}[t]
\centering
\caption{ {RMSE of matrix completion with different ranks.}}
\label{tab:MF high-rank}
    \begin{tabular}{|c|c|c|c|c|c|c|}
      \hline
      \diagbox[width=\dimexpr \textwidth/12+4\tabcolsep\relax, height=0.5cm]{ Methods }{Rank}
                  &5       & 10     & 20     & 40     & 80    & 100   \\ \hline
MLAM              &\textbf{0.089}   & \textbf{0.085}  & \textbf{0.078}  & \textbf{0.072}  & \textbf{0.053} & \textbf{0.052} \\ \hline
SGD               &0.292   & 0.405  & 0.689  & $>$1  & $>$1   & $>$1   \\ \hline
ALS               &0.864   & 0.942  & $>$1  & $>$1  & $>$1   & $>$1   \\ \hline
 {DMF}              & 0.228  & 0.263  & 0.392  & 0.643 & $>$1 & $>$1   \\ \hline
 {UMF}              & 0.363  & 0.376  & 0.417  & 0.582 & $>$1 & $>$1  \\ \hline
\end{tabular}
\end{table}

\begin{table}[t]
\centering
\caption{ {RMSE of rank-10 matrix completion with variant factorized matrix dimension $p$.}}
\label{tab:MF k}
    \begin{tabular}{|c|c|c|c|c|c|}
      \hline
      \diagbox[width=\dimexpr \textwidth/8+4\tabcolsep\relax, height=0.5cm]{ Methods }{$p$}
                   & 10   & 20   & 40   & 80   & 100  \\ \hline
MLAM             & \textbf{0.09}  & \textbf{0.12}  & \textbf{0.19}  & \textbf{0.30}  & \textbf{0.36} \\ \hline
SGD              & 0.31  & 0.78  & $>$1  & $>$1  & $>$1  \\ \hline
ALS              & 0.89  & $>$1  & $>$1  & $>$1  & $>$1  \\ \hline
 {DMF}              & 0.26  & 0.53  & 0.81  & $>$1  & $>$1  \\ \hline
 {UMF}              & 0.37  & 0.68  & 0.87  & $>$1  & $>$1 \\ \hline
\end{tabular}
\end{table}

\begin{table}[t]
\centering
\caption{ {Time complexity of different methods.}}
\label{tab:time}
    \begin{tabular}{|c|c|c|c|c|c|}
      \hline
     Methods      & MLAM   & DMF   & UMF   & ALS   & SGD   \\ \hline
     Time(ms)     & 4618 &  3704& 2243 & \textbf{1272} & 1518 \\ 
 \hline
\end{tabular}
\end{table}

\begin{table*}[t]
\centering
\caption{RMSEs of mixed matrix completion with different choice of $p$.}
\label{tab:MF mix}
    \begin{tabular}{|c|c|c|c|c|c|c|c|c|c|c|c}
      \hline
      \diagbox[width=\dimexpr \textwidth/8+4\tabcolsep\relax, height=0.5cm]{ $p$ }{Rank}
                   & 10   & 20   & 30   & 40   & 50  & 60   & 70   & 80   & 90  & 100  \\ \hline
10              &0.120  &0.081  &0.065  &0.055  &0.049 &0.046  & 0.043 & 0.042 & 0.042 & \textbf{0.039}  \\ \hline
50              & 0.250  & 0.130  & 0.095  & 0.075  & 0.067 & 0.055  & 0.052  & 0.048  & 0.046  & \textbf{0.045} \\ \hline
100              & 0.450  & 0.210  & 0.140 & 0.100  & 0.081 & 0.068  & 0.061 & 0.056  & 0.052  & \textbf{0.048} \\ \hline
\end{tabular}
\end{table*}

\subsubsection{Blind matrix completion}
In this part, we consider more challenging cases where the rank information is not known. We consider these are blind matrix completion problems which are typically intractable using previous methods. We first test our MLAM method and standard methods in the case where a set of matrices completion problems have the same but unknown rank. Then we will further test our MLAM method by  more difficult case which has a set of matrices completion problems with different unknown ranks. 

In the first case, we test our MLAM method and two standard methods on a set of rank-10 matrix completion problems with variant factorized matrix dimension $p$. In Table \ref{tab:MF k} we report the results of applying different $p$ for reconstructing a rank-10 matrix through the five tested approaches. We can see that the proposed MLAM method is more robust when there is a mismatch between $p$ and the true rank compared to the other methods.

 {When we take a large gap such as $p=80$ or $p=100$ to reconstruct a rank-10 matrix, the MLAM method still achieves acceptable performances. Meanwhile, the rest of the tested methods quickly degrade with the increase of the mismatch between $p$ and the rank values. The RMSEs of the existing methods quickly arise to more than  $100\%$ when $p\geq40$. In contrast, our MLAM method shows a good tolerance on the increase of the difference between the real rank and $p$. Thus, the MLAM method  does not require an accurate rank information to achieve a successful} matrix completion. 
% \begin{figure}[t]
%     \centering
%     \includegraphics[width=0.4\textwidth]{Fig/K_chocies_v2.png}
%     \caption{Averaged RMSE curves of three methods over 100 testing samples on variant factorized matrix dimension $p$.}
%     \label{fig:p}
% \end{figure}

\begin{table}[t]
\centering
\caption{High dimensional GMM simulation results.}
\label{tab:GMM_dimension}
    \begin{tabular}{|c|c|c|c|c|c|}
      \hline
      \diagbox[width=\dimexpr \textwidth/8+4\tabcolsep\relax, height=0.5cm]{ Methods }{Dimension}
                   & 4   & 8   & 16   & 32   & 64  \\ \hline
EM              & 6.52  & 12.64  & 23.83  & 47.10  & 96.88\\ \hline
MLAM              & \textbf{6.44}  & \textbf{12.42}  & \textbf{23.05} & \textbf{46.93}  & \textbf{95.72}  \\ \hline
\end{tabular}
\end{table}
\begin{figure}[t]
    \centering
    \includegraphics[width=0.3\textwidth]{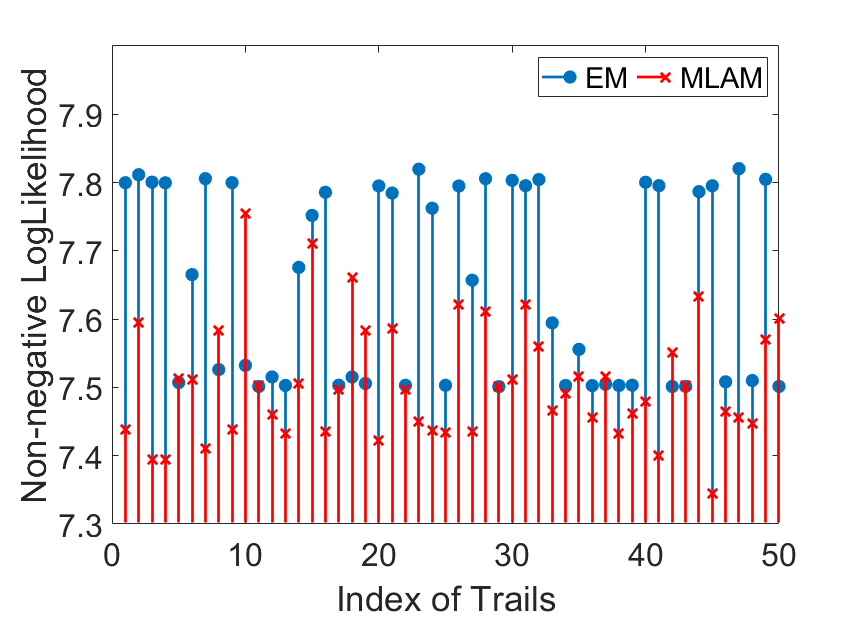}
    \caption{The performance of EM and MLAM over 50 testing samples. Averaged non-negative log-likelihood of 50 trails through EM algorithm is 7.77, while the result through MLAM model is 7.56.}
    \label{fig:GMM_2_50list}
\end{figure}
\begin{figure}[t]
	\centering
	\hspace*{\fill}
	\subfigure[EM clustering result.]{
% 		\label{fig:GMM_2_50list} %
		\includegraphics[width=0.2\textwidth]{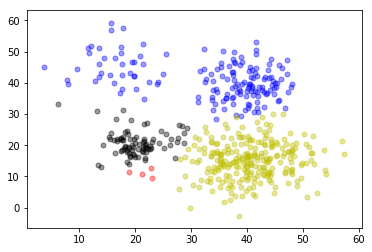}}
		\hfill
	\subfigure[MLAM clustering result.]{
% 		\label{fig:GMM_2_OneSample} %% 
		\includegraphics[width=0.2\textwidth]{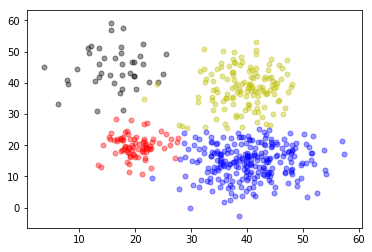}}
	\hspace*{\fill}
	\caption{Clustering results of one GMM problem with random initialization.}
    \label{fig:GMM_2_OneSample}
\end{figure}
\begin{figure*}[h]
    \centering
    \includegraphics[width=0.65\textwidth]{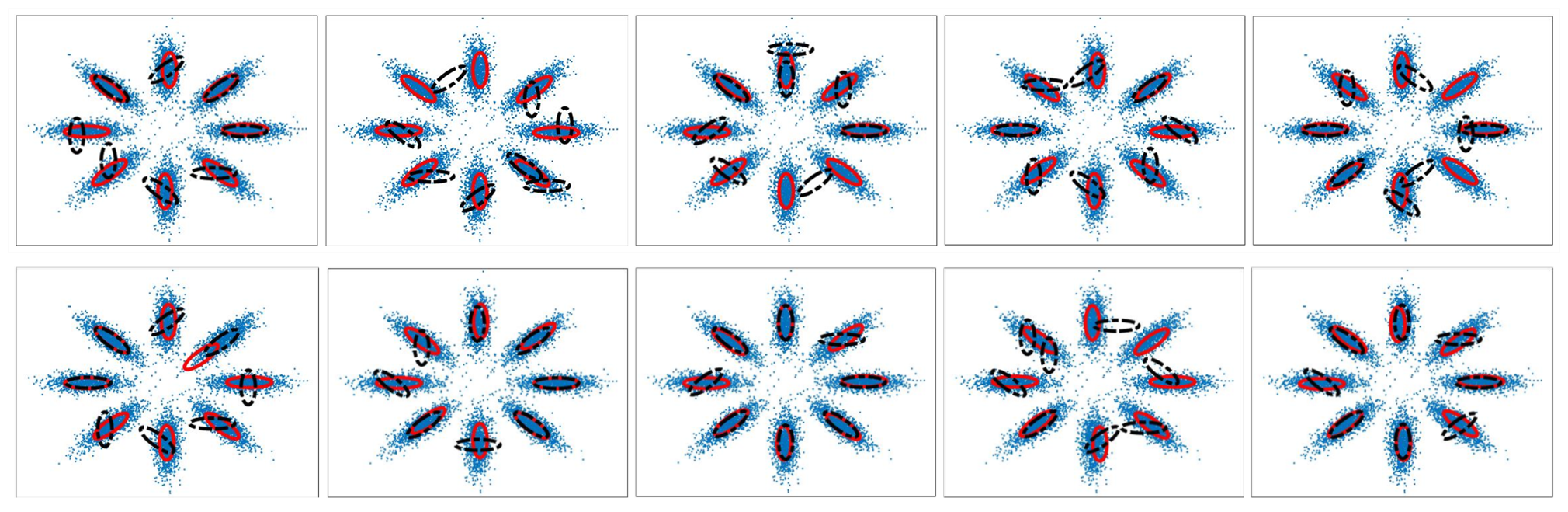}
    \caption{EM and MLAM performance on 10 flower-shaped data. Black and red dash lines denote EM and MLAM clustering results, respectively}
    \label{fig:Flower}
\end{figure*} 

% \subsubsection{Scenario 5: Mixed rank samples}
Then, we test our MLAM method on matrix completion problems with different ranks for the matrix $\bm{R}$. This means that in the training and testing sample sets, there are mixed matrices with different ranks. Thus the learned algorithm is required to adapt on matrix completion problems across different ranks.

In Table \ref{tab:MF mix}, RMSE results of three sets of matrix completion problems are listed. For each set, $100$ training samples and $100$ testing samples are generated by setting their ground truth rank uniformly distributed between 10 and 100 with stepsize 10. We take three different choices of $p$ for each set, and calculate the mean RMSE on the samples of each rank respectively.

The proposed MLAM method in the three cases performs generally well on samples with different ranks. The comparison of the three cases further reveals that our method of choosing $p=10$ achieves the best performance in all samples, especially on low-rank scenarios where it provides significantly better results than the others. For the other two choices of $p$, our method also has comparable performances on high-rank samples, and achieves sufficiently good accuracy on the majority of the samples. 

 {The major computational cost in each iteration of MLAM is from the network-based computation of the gradient update term at inner loops. In this instance, due to the implementation of the networks at each inner loop iteration, the computational complexity of the MLAM approach is significantly regarded to the number of the inner and outer loops.    
 
% In this stage, we admit that the proposed MLAM is relatively more time-consuming than the existing methods, and it is expected to be further improved on the aspect of the computational speed.
In the future work, we will further improve the computational speed of the MLAM to reduce the time cost for better practical applications. 
} 

In summary, we have been shown that our MLAM method is capable of solving the matrix completion problem without any prior information. Typically, these problems have underlying complicated landscapes geometries. Therefore, it is hard for standard gradient descent based methods to achieve satisfactory results. The proposed MLAM model makes it possible to find good solutions through learning an appropriate updating rule that is dynamic and adaptive.

\subsection{Numerical Results on Gaussian Mixture Model}
In this section, we apply  MLAM method to  GMM problems. Given the data set $\bm{X}=\{\bm{x}_g\}_{g=1}^G$ ($G$ denotes the number of observation samples), we optimize the mean cluster centre $\bm{\mu}_k$ and mixing proportion $\pi_k$ whilst keeping the covariance $\mathbf{\Sigma}_k$ frozen, which is similar to the optimization of the \textit{k-means} problem. We consider one GMM problem as one sample in the training and testing set. For solving each GMM problem, we set the maximum alternating steps as $100$, and record the negative log-likelihood $F_{ne}$ trajectory according to  (\ref{Ne-ML-GMM}). Similar to the settings in the matrix completion simulation, we also choose $t_{in}=10$ and $t_{out}=10$ for all the simulation scenarios in the GMM problems.

EM algorithm has been adopted for comparison. The stopping criterion for EM algorithm is $\arrowvert F_{ne}(t-1)-F_{ne}(t)\arrowvert<10^{-4}$. EM algorithm and our MLAM method all start from the same random initialization for $\bm{\mu}_k$ and $\pi_k$. 

% \subsubsection{Different dimensional GMM}
We first start from the 2-dimensional GMM problems, whose data vector $\bm{x}_g\in\mathbb{R}^{2}$. 
Given four clusters ($K=4$), $G=500$ data points in $\bm{X}=\{\bm{x}_g\}_{g=1}^G$, we show $50$ tested results with random initialization in Fig \ref{fig:GMM_2_50list}. We only provide 500 data points here to increase the difficulty for optimizing these GMM problems due to the limited number of observations. From Fig \ref{fig:GMM_2_50list}, it can be seen that on these 50 tests, the MLAM method outperforms the EM algorithm in most cases. The mean negative log-likelihood $F_{ne}$ of MLAM method on these $50$ samples is 7.52 while that of the EM algorithm is 7.75. More specifically, we randomly select one clustering result among these 50 trails which is shown in Fig.\ref{fig:GMM_2_OneSample}, which clearly shows that the MLAM method obtains much better clustering results than the EM algorithm.

Furthermore, we conduct simulations on a flower-shaped synthetic data (G = 10,000)
with random initializations. Each cluster in the flower-shaped data is composed by Gaussian-distributed samples. This typically is a hard problem as the anisotropic clusters lead to extensive local optimum. 10 randomly selected optimization can be found
in Fig \ref{fig:Flower}, in which our algorithm consistently achieves nearly optimal clustering but the results from EM are
highly biased. Although being illustrative in $2$ dimensions, the data in Fig. \ref{fig:Flower} consists of $8$ anisotropic clusters, which might be the main reason that the EM method converges to bad local optimum. As has been proved in \cite{jin2016local}, when solving GMMs with more than $2$ clusters ($K>2$), the EM method is highly likely to converge to arbitrarily bad local optimum under random initialization. More specifically, The EM method is a special case of Beyasian variational inference, and provides a tractable solver to maximize the lower bound of the log-likelihood of GMM problems. More importantly, maximizing the log-likelihood is equivalent of using the Kullback-Leibler (KL) divergence in estimating two distributions \cite{xu1996convergence}. Thus, the performance of the EM method is basically limited to the intrinsic nature of the KL divergence, which can output infinitely large values with gradient vanishing issues when two distributions are well-separated \cite{xu2016global,jin2016local}. In other words, the EM method is highly sensitive to initialization and may converge to bad local optimum from random initialization, the phenomenon that has been studied in \cite{li2019general}. 
In contrast, the proposed MLAM method consistently achieves the global estimation, verifying the global optimization nature of our method. Therefore, our MLAM method obtains significant improvements on accuracy, even for some challenging scenarios such as insufficient observations and anisotropic clusters in these illustrative evaluations.

% \subsubsection{High dimensional GMM }
We further evaluate our method in estimating high dimensional GMM problems. Several sets of high dimensional synthetic data ($G=500$, $K=4$) are also randomly generated for the evaluation, with dimensions $4, 8, 16, 32$ and $64$. The averaged negative log-likelihoods of $100$ tests for each testing dimension are reported in Table \ref{tab:GMM_dimension}. It is clear that from Table \ref{tab:GMM_dimension} our MLAM method outperforms the EM algorithm on all the high dimensional sample sets and the variation on dimensions does not affect the performance of our method, while this typically decreases the performance of EM algorithm in general.

In this stage, it has been verified that our proposed method is able to solve non-convex multi-variable problem with non-linear relationships between variables and observation data. Even without closed-from landscapes in these problems, our MLAM method still successfully finds good solutions whilst EM algorithm fails.

\section{Conclusion}
We have proposed a meta-learning based alternating minimization (MLAM) method for solving non-convex optimization problems. The learned algorithm has been verified to have a faster convergence speed and better performances than existing alternating minimization (AM)-based methods. To achieve that, our MLAM method has employed LSTM-based meta-learners to build an interaction between variable updates and the global loss. In this way, the variables are updated by the LSTM networks with frozen parameters at inner loops, which are then updated by minimizing accumulated global losses at outer loops. This paper is just a proof of the concept that a less greedy and learning-based solution for non-convex problem could surpass both of the traditional model-based and learning-based methods. More importantly, these concepts: optimize each independent solution step in-exhaustively by globally learning the optimization strategy, and integrate these independent steps by the bi-level meta-learning optimization model, spark a new direction of improving the optimization-inspired solutions for advanced performance. It reveals that applying neural network based behavior to assist the well-established algorithmic principle, instead of replacing it by black-box network behavior, could bring significant advances. 

For the future work, we plan to apply the proposed MLAM method to more challenging and practical non-convex problems, such as dictionary learning in compress sensing, blind image super resolution, and MIMO beamforming. We would also like to extend some theoretical analysis on our MLAM model.
Moreover, the MLAM can be directly apply to solve non-convex problem online without pre-training. It is deserved to discover the application of applying MLAM as an algorithm without training for non-convex problems.

\ifCLASSOPTIONcaptionsoff
  \newpage
\fi



%\bibliographystyle{IEEEtran}
%\begin{thebibliography}{1}

%\end{thebibliography}
\bibliographystyle{IEEEtran}
\addcontentsline{toc}{section}{\refname}\bibliography{Bib_TNNLS}

% \begin{comment}

\begin{IEEEbiography}[{\includegraphics[width=1in,height=1.10in,clip,keepaspectratio]{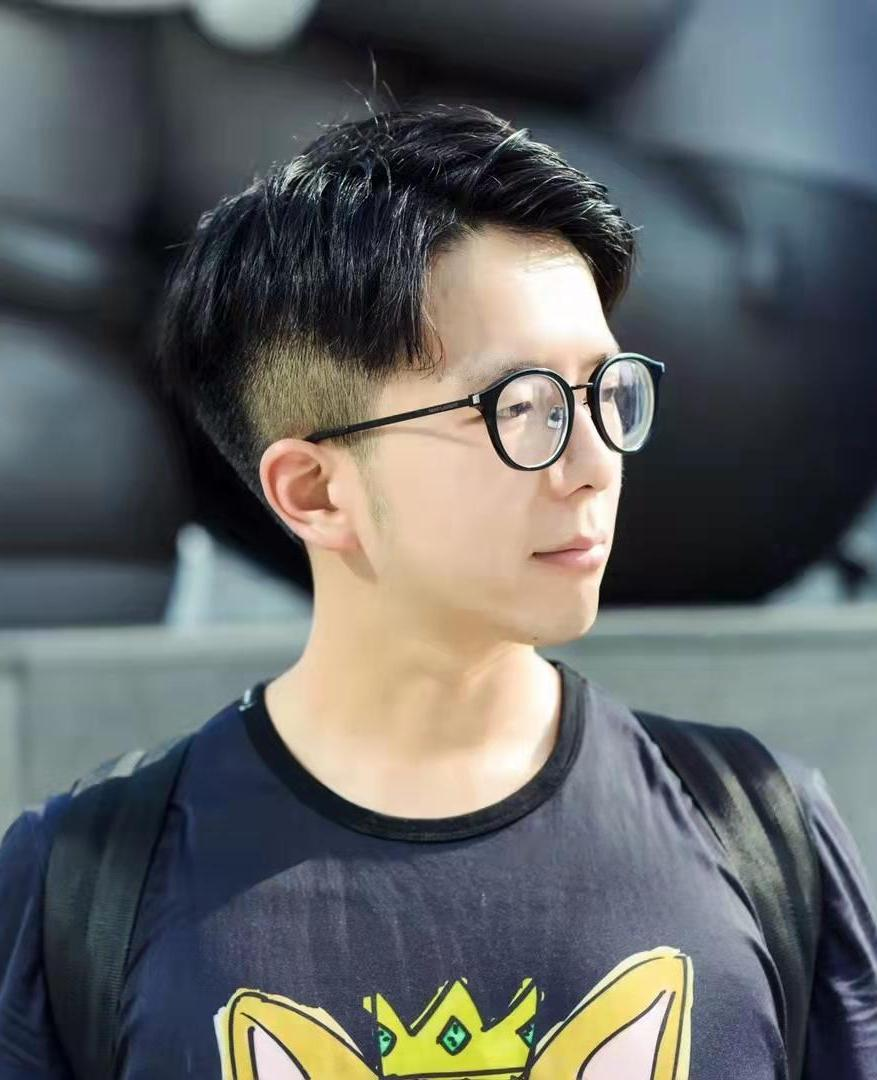}}]
{Jingyuan Xia} received his B.Sc. and M.Sc. degree from the National University of Defense Technology (NUDT), Changsha, China, and received his Ph.D. degree from Imperial College London (ICL) in 2020. He has been a lecturer with the college of the Electronic Science at the National University of Defense Technology (NUDT) since 2020. His current research interests include machine learning, non-convex optimization, and representation learning.
\end{IEEEbiography}

\begin{IEEEbiography}[{\includegraphics[width=1in,height=1.10in,clip,keepaspectratio]{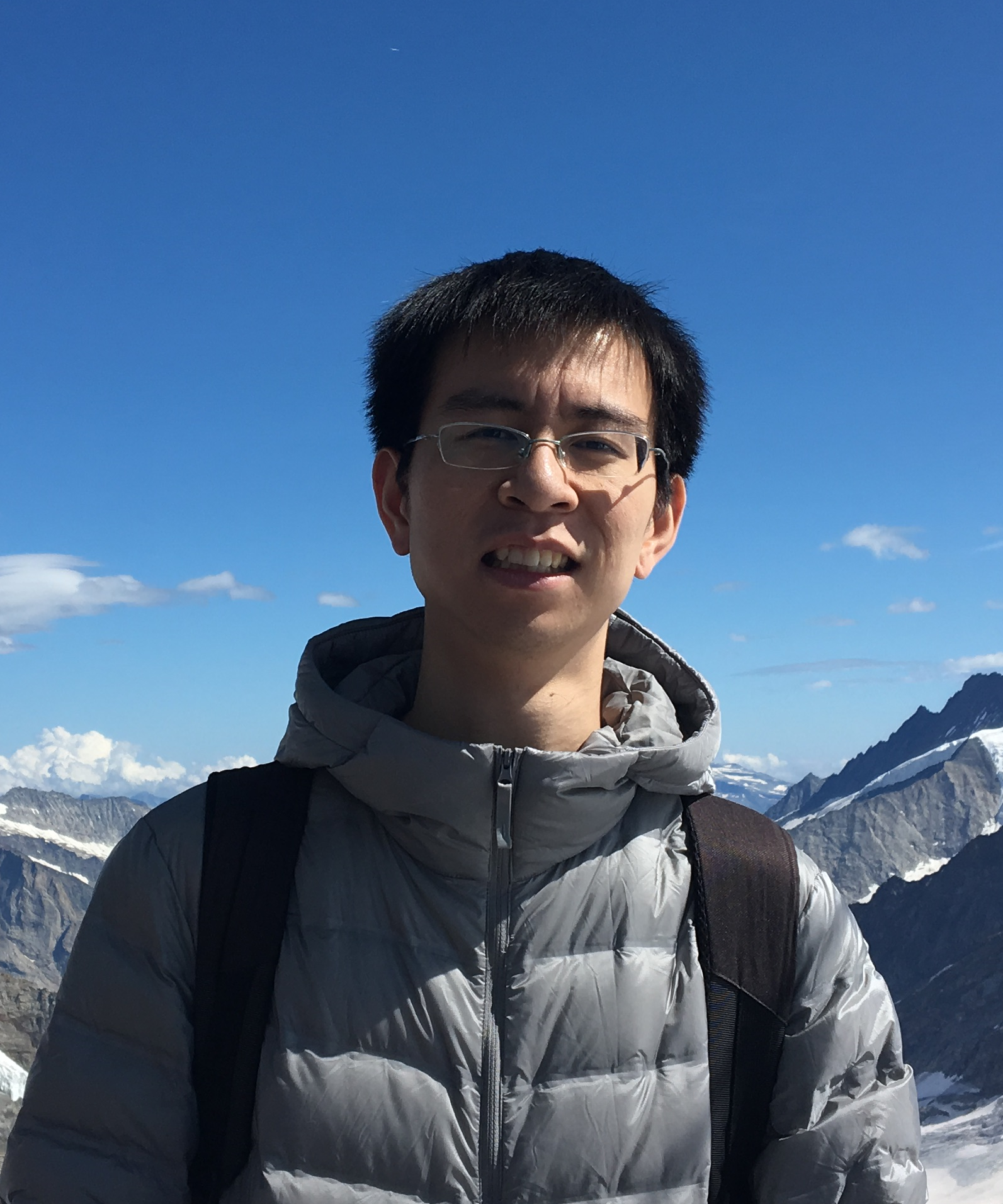}}]
{Shengxi Li} (S’14-M’22) received his Bachelor degree in Beihang University in Jul. 2014, the Master degree in Mar. 2016, also in Beihang University, and got his PhD degree in Aug. 2021, in EEE department of Imperial College London, under the supervision by Prof. Danilo Mandic. He is now an associate professor in Beihang University. His research interests include generative models, statistical signal processing, rate distortion theory, and perceptual video coding. He is the receipts of Imperial Lee Family Scholarship and Chinese Government Award for Outstanding Self-financed Students Abroad.
\end{IEEEbiography}

\begin{IEEEbiography}[{\includegraphics[width=1in,height=1.10in,clip,keepaspectratio]{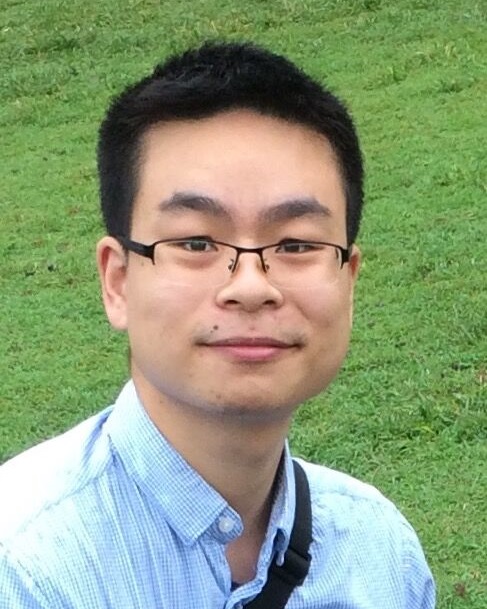}}]
{Jun-Jie Huang} (Member, IEEE) received the B.Eng. (Hons.) degree with First Class Honours in Electronic Engineering and the M.Phil. degree in Electronic and Information Engineering from The Hong Kong Polytechnic University, Hong Kong, China, in 2013 and 2015, respectively, and the Ph.D. degree from Imperial College London (ICL), London, U.K., in 2019. He is a Lecturer with College of Computer Science, National University of Defence Technology (NUDT), Changsha, China. During 2019 - 2021, he was a Postdoc with Communications and Signal Processing (CSP) Group, Electrical and Electronic Engineering Department, ICL, London, U.K. His research interests include the areas of computer vision, signal processing and deep learning.
\end{IEEEbiography}

\begin{IEEEbiography}[{\includegraphics[width=1in,height=1.10in,clip,keepaspectratio]{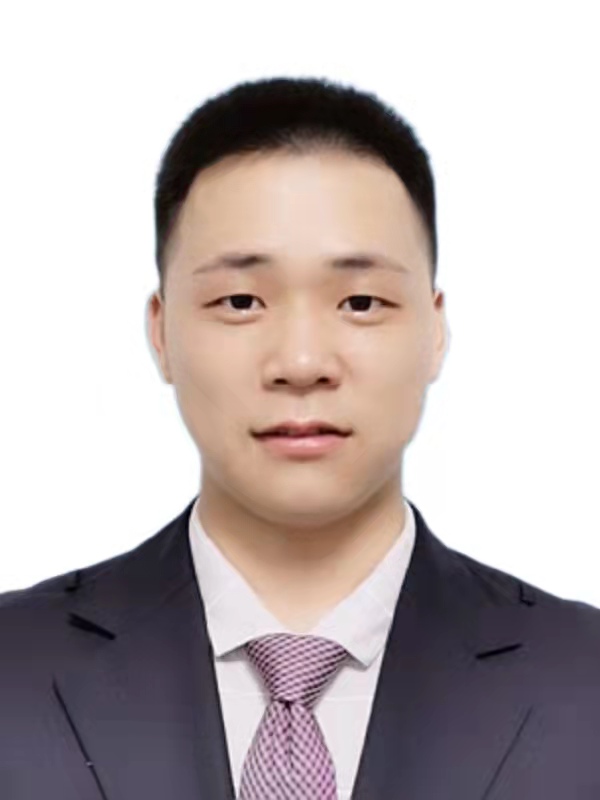}}]
{Zhixiong Yang} received the B.Sc. degree from the Northeastern University, China. He is now pursuing a M.Sc. degree at the College of Electronic Science, the National University of Defense Technology. His research interests include deep learning on signal processing and image processing.
\end{IEEEbiography}

\begin{IEEEbiography}[{\includegraphics[width=1in,height=1.10in,clip,keepaspectratio]{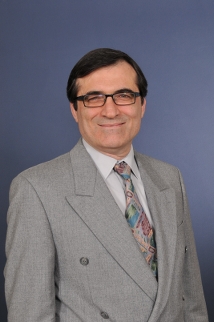}}]
{Imad M. Jaimoukha} received the B.Sc degree in Electrical Engineering from the University of Southampton, UK, in 1983, and the MSc and PhD degrees in control systems from Imperial College London, UK, in 1986 and 1990, respectively. 
He was a Research Fellow with the Centre for Process Systems Engineering at Imperial College London from 1990 to 1994. Since 1994, he has been with the Department of Electrical and Electronic Engineering, Imperial College London. His research interests include robust and fault tolerant control, system approximation, and global optimization.
\end{IEEEbiography}

\begin{IEEEbiography}[{\includegraphics[width=1in,height=1.10in,clip,keepaspectratio]{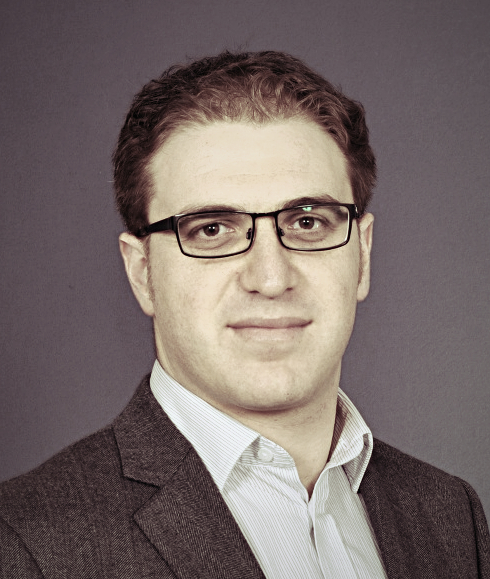}}]
{Deniz Gündüz} received his M.S. and Ph.D. degrees from NYU Tandon School of Engineering (formerly Polytechnic University) in 2004 and 2007, respectively. After his PhD, he served as a postdoctoral research associate at Princeton University, as a consulting assistant professor at Stanford University, and as a research associate at CTTC in Barcelona, Spain. ln Sep. 2012, he joined the Electrical and Electronic Engineering Department of Imperial College London, UK, where he is currently a Professor of Information Processing, and serves as the deputy head of the Intelligent Systems and Networks Group. He is also a part-time faculty member at the University of Modena and Reggio Emilia, Italy, and has held visiting positions at University of Padova (2018-2020) and Princeton University (2009-2012). Dr. Gündüz is a Fellow of the IEEE, and a Distinguished Lecturer for the IEEE Information Theory Society (2020-22). He is the recipient of a Consolidator Grant of the European Research Council (ERC) in 2022, the IEEE Communications Society - Communication Theory Technical Committee (CTTC) Early Achievement Award in 2017, a Starting Grant of the European Research Council (ERC) in 2016, and several best paper awards. 
\end{IEEEbiography}

% \begin{IEEEbiography}[{\includegraphics[width=1in,height=1.10in,clip,keepaspectratio]{AuthorFigures/XinwangLiu2.jpg}}]
% {Xinwang Liu} received his PhD degree from National University of Defense Technology (NUDT), China. He is now full Professor of School of Computer, NUDT. His current research interests include kernel learning and unsupervised feature learning. Dr. Liu has published 60+ peer-reviewed papers, including those in highly regarded journals and conferences such as IEEE T-PAMI, T-KDE, T-IP, T-NNLS, T-MM, T-IFS, NeurIPS, ICCV, CVPR, AAAI, IJCAI, etc.
% \end{IEEEbiography}    

% \end{comment}

% \end{IEEEbiography}


%\newpage



%\vfill


\end{document}